%% file: main.tex
\documentclass[a4,11pt]{article}

\usepackage{geometry, fancybox, framed}
\usepackage{enumerate}
\usepackage{amsmath,amssymb, bm}
\usepackage{mathtools}
\usepackage{graphics}
\usepackage{multirow}

\graphicspath{{fig/}}
\allowdisplaybreaks[4]

\usepackage{my-style}

\begin{document}
\title{Dataset-Free Weight-Initialization on Restricted Boltzmann Machine}
\author{
\large{Muneki Yasuda}\thanks{Graduate School of Science and Engineering, Yamagata University} 
\and \large{Ryosuke Maeno}\thanks{Techno Provide Inc.}
\and \large{Chako Takahashi}\footnotemark[1]
}
\date{}
\maketitle

\begin{abstract}
In feed-forward neural networks, dataset-free weight-initialization methods such as LeCun, Xavier (or Glorot), and He initializations have been developed. 
These methods randomly determine the initial values of weight parameters based on specific distributions (e.g., Gaussian or uniform distributions) 
without using training datasets. 
To the best of the authors' knowledge, such a dataset-free weight-initialization method is yet to be developed 
for restricted Boltzmann machines (RBMs), which are probabilistic neural networks consisting of two layers. 
In this study, we derive a dataset-free weight-initialization method for Bernoulli--Bernoulli RBMs based on statistical mechanical analysis. 
In the proposed weight-initialization method, the weight parameters are drawn from a Gaussian distribution with zero mean. 
The standard deviation of the Gaussian distribution is optimized based on our hypothesis 
that a standard deviation providing a larger layer correlation (LC) between the two layers improves the learning efficiency. 
The expression of the LC is derived based on a statistical mechanical analysis. 
The optimal value of the standard deviation corresponds to the maximum point of the LC.
The proposed weight-initialization method is identical to Xavier initialization in a specific case (i.e., when the sizes of the two layers are the same, 
the random variables of the layers are $\{-1,1\}$-binary, and all bias parameters are zero). 
The validity of the proposed weight-initialization method is demonstrated in numerical experiments using a toy and real-world datasets.
\end{abstract}

\section{Introduction}
\label{sec:introduction}

A restricted Boltzmann machine (RBM) is a probabilistic neural network defined on a bipartite undirected graph 
consisting of two layers: visible and hidden layers~\cite{Smolensky1986,Hinton2002}. 
The visible layer consists of visible variables that directly correspond to the data points, 
whereas the hidden layer comprising hidden variables does not, wherein both visible and hidden variables are random variables. 
The hidden layer creates complex correlations among the visible variables. 
The RBMs have several applications, such as collaborating filtering~\cite{Salakhutdinov2007}, dimensionality reduction~\cite{Hinton2006b,Yasuda2023}, 
classification~\cite{Larochelle2012,Yokoyama2019}, anomaly detection~\cite{Fiore2013,Sekimoto2024}, and deep learning~\cite{Hinton2006,DBM2009,Kanno2021}. 
Further, the RBMs have been actively investigated in the fields of statistical mechanics~\cite{Barra2011,Barra2017,Hartnett2018,Decelle2021}. 

Learning parameters in gradient-based successive learning are initialized using a specific approach, which can affect the learning result.
Therefore, selecting an appropriate initialization approach is essential. 
Finding a universal optimal-initialization approach can be challenging 
because the optimal initialization strongly depends on a given dataset, 
and ultimately, the optimal initial values are the learning solution. 
An initialization method that is dataset-free and stably provides good learning solutions is useful. 
Such useful weight-initialization methods exist for (non probabilistic) feed-forward neural networks, 
e.g., LeCun~\cite{LeCun1998}, Xavier (or Glorot)~\cite{Glorot2010}, and He~\cite{He2015} initializations, 
Herein the weight parameters of the feed-forward neural network are initialized by small random values drawn from a specific distribution 
(and its bias parameters are usually initialized to zero). 
These three weight-initialization methods depend only on the structure of the network. 
However, such a weight-initialization method does not exist in RBM learning.

This study investigates a dataset-free weight-initialization method for Bernoulli--Bernoulli RBMs 
whose visible variables are $\{-1,1\}$-binary and hidden variables are $\{0,1\}$- or $\{-1,1\}$-binary. 
In the proposed weight-initialization method, the bias parameters are initialized by a fixed value 
and the weight parameters are initialized by small random values drawn from a specific distribution, 
similar to the aforementioned weight-initialization methods for feed-forward neural networks. 
Wherein the distribution generating initial weights is assumed to be a Gaussian distribution with zero mean and standard deviation $\sigma$. 
The value of $\sigma$ is determined by a criterion based on the hypothesis 
that a $\sigma$-value providing a larger layer correlation (LC) between the visible and hidden layers (i.e., a statistical average of covariance between the two layers) 
improves the learning efficiency. 
The learning efficiency mentioned here is considered as the growth rate of the training likelihood.

In section \ref{sec:statistical_mechanical_analysis}, the LC is evaluated based on a statistical mechanical approach, i.e., a mean-field analysis, 
similar to that reported in references~\cite{Barra2011,Barra2017}, 
and the criterion for the determination of $\sigma$ is obtained from the evaluation result. 
The proposed weight-initialization method is identical to Xavier initialization~\cite{Glorot2010} in a specific case.
In section \ref{sec:experiment}, the proposed weight-initialization method is applied to RBM learning experiments using a toy and real-world datasets, 
and the obtained numerical results support the validity of the proposed weight-initialization method. 
Section \ref{sec:summary} summarizes this study and discusses the future scope.

\section{Restricted Boltzmann Machine}
\label{sec:RBM}

Consider an RBM in which the visible and hidden layers consist of $n$ visible and $m$ hidden variables, $\bm{v} := \{v_i \mid i \in V:=\{1,2,\ldots, n\}$\} 
and $\bm{h} := \{h_j   \mid j \in H:=\{1,2,\ldots, m\}\}$, respectively. 
The visible and hidden variables are supposed to be binary discrete: $v_i \in \mcal{I}:= \{-1,1\}$ 
and $h_j \in \mcal{X}_{\mrm{h}}$, where $\mcal{X}_{\mrm{h}} = \mcal{B}:=\{0,1\}$ or $\mcal{X}_{\mrm{h}}=\mcal{I}$, 
i.e., this RBM is a Bernoulli--Bernoulli RBM. 
The RBM is expressed as
\begin{align}
P(\bm{v}, \bm{h} \mid \theta) := \frac{1}{Z(\theta)} \exp\Big(\sum_{i \in V} b_i v_i +\sum_{j \in H}c_j h_j + \sum_{i\in V}\sum_{j\in H}w_{i,j}v_ih_j \Big),
\label{eqn:RBM}
\end{align}
where $Z(\theta)$ represents the partition function defined as
\begin{align*}
Z(\theta):= \sum_{\bm{v}} \sum_{\bm{h}} \exp\Big(\sum_{i \in V} b_i v_i +\sum_{j \in H}c_j h_j + \sum_{i\in V}\sum_{j\in H}w_{i,j}v_ih_j \Big),
\end{align*}
where $\sum_{\bm{v}}$ and $\sum_{\bm{h}}$ denote multiple summations over all possible configurations of assigned variables, 
e.g., $\sum_{\bm{v}} = \sum_{v_1 \in \mcal{I}}\sum_{v_2 \in \mcal{I}} \cdots \sum_{v_n \in \mcal{I}}$.
Here, $\bm{b}:=\{b_i \mid i \in V\}$ and $\bm{c} = \{c_j \mid j \in H\}$ are the biases for visible and hidden variables, respectively, 
and $\bm{w}:=\{w_{i,j} \mid i \in V,\> j\in H\}$ are the weights between the visible and hidden layers. 
These learning parameters are collectively denoted by $\theta$. 
For a give dataset consisting of $N$ data points, $D:= \{\mbf{v}^{(\mu)} \mid \mu = 1,2,\ldots, N\}$ where 
$\mbf{v}^{(\mu)}:=\{\mrm{v}_i^{(\mu)} \in \mcal{I} \mid i \in V\}$, 
the log likelihood is defined as
\begin{align}
L(\theta) := \frac{1}{N}\sum_{\mu=1}^N \ln P(\mbf{v}^{(\mu)} \mid \theta),
\label{eqn:likelihood}
\end{align}
where $P(\bm{v}\mid \theta)$ is the marginal distribution of RBM: $P(\bm{v}\mid \theta) = \sum_{\bm{h}}P(\bm{v}, \bm{h}\mid \theta)$.
The RBM learning is archived by the maximization of the log likelihood with respect to the learning parameters. 
Usually, the maximization is performed based on a gradient-based method. 
However, the gradients of the log likelihood include the model expectations of the RBM, 
whose evaluation is computationally infeasible when the size of the RBM is large. 
Practical approximate learning methods have been proposed, e.g., contrastive divergence (CD)~\cite{Hinton2002,Tieleman2008}, 
parallel tempering~\cite{Desjardins2010}, and spatial Monte Carlo integration~\cite{Selkimoto2023}.

The learning parameters in the RBM, $\theta$, are initialized at the beginning of the RBM learning. 
In this study, we consider an initialization: the visible and hidden biases are initialized to constant values, $b_i = b$ and $c_j = c$, 
and the weights are initialized by small random values generated from a distribution $P_{\mrm{ini}}(\bm{w})$. 
The initial value of the visible biases, $b$, is fixed to zero, and that of the hidden biases, $c$, is set to zero when $\mcal{X}_{\mrm{h}} = \mcal{I}$, 
and to zero or a negative value when $\mcal{X}_{\mrm{h}} = \mcal{B}$. 
The negative value initialization encourages sparse representations of the hidden layer~\cite{Hinton2012}.
Therefore, the initial RBM is expressed as
\begin{align}
P(\bm{v}, \bm{h} \mid \theta_{\mrm{ini}})\propto \exp\Big(b\sum_{i \in V} v_i + c\sum_{j \in H}h_j + \sum_{i\in V}\sum_{j\in H}w_{i,j}v_ih_j\Big),
\label{eqn:initial_RBM}
\end{align}
where $\bm{w}$ are the initial weights drawn from $P_{\mrm{ini}}(\bm{w})$. 
In the initial RBM, $b$ should vanish because it is fixed to zero in our initialization. 
However, we leave $b$ in this expression for the convenience of the subsequent analysis.
In this study, $\bm{w}$ are independently drawn from a Gaussian distribution with zero mean 
and standard deviation $\sigma := \beta /\sqrt{n + m}$: 
\begin{align*}
P_{\mrm{ini}}(\bm{w})=P_{\mrm{ini}}(\bm{w} \mid \beta) 
= \prod_{i \in V}\prod_{j \in H} \sqrt{\frac{n + m}{2 \pi \beta^2}}\exp\Big(- \frac{(n + m)}{2 \beta^2 }w_{i,j}^2\Big),
\end{align*}
where $\beta$ is a nonnegative value. 
When $\mcal{X}_{\mrm{h}} = \mcal{I}$, the initial RBM corresponds to a bipartite Sherrington--Kirkpatrick (SK) model~\cite{Hartnett2018}.

This study aims to find the value of $\beta$ that can increase learning efficiency. 
The learning efficiency mentioned here is considered as the growth rate of the training likelihood.
We hypothesize that the distribution with a large LC between the visible and hidden layers is more efficient. 
The information of one layer is easily transferred to the other layer when the LC is large.
The smooth transfer of information in the initial state increases the learning efficiency at least in the early stage of learning. 
In the initial RBM presented in equation (\ref{eqn:initial_RBM}), we consider a statistical average of the sum-of-covariances defined by
\begin{align}
\chi(\beta)\propto\sum_{i \in V}\sum_{j \in H}\int_{-\infty}^{+\infty}d\bm{w}\, P_{\mrm{ini}}(\bm{w} \mid \beta) \Big(\mathbb{E}_{\mrm{ini}}[v_i h_j] - 
\mathbb{E}_{\mrm{ini}}[v_i]\mathbb{E}_{\mrm{ini}}[h_j]\Big),
\label{eqn:layer_correlation}
\end{align}
as the LC, where $\mathbb{E}_{\mrm{ini}}[\cdots]$ represents the expectation on the initial RBM.
The LC in equation (\ref{eqn:layer_correlation}) includes the covariances between the visible and hidden variables. 
In our hypothesis, the $\beta$ value that maximizes $|\chi(\beta)|$,
\begin{align}
\beta_{\mrm{max}} := \argmax_{\beta} |\chi(\beta)|,
\label{eqn:def_beta_max}
\end{align}
would be the most efficient. 
Therefore, the proposed initialization is summarized as follows: 
the biases $\bm{b}$ and $\bm{c}$ are initialized to constant values, i.e., 
$b_i = 0$ and $c_j =0$ when $\mcal{X}_{\mrm{h}} = \mcal{I}$ 
and $b_i = 0$ and $c_j =c$, where $c \leq 0$, when $\mcal{X}_{\mrm{h}} = \mcal{B}$; 
further, the weights $\bm{w}$ are initialized by small random values independently drawn from a Gaussian distribution with zero mean 
and $\sigma = \beta_{\mrm{max}} /\sqrt{n + m}$. 
Notably, when $c = 0$ and $\beta_{\mrm{max}} = \sqrt{2}$, 
the proposed weight initialization is identical to that of the Xavier (Glorot) initialization~\cite{Glorot2010}. 

An interpretation of the LC from a statistical mechanical point of view is as follows.
Here, we define a (statistical-averaged) free energy as
\begin{align}
f(\beta):=-\frac{1}{n + m} \int_{-\infty}^{+\infty}d\bm{w}\, P_{\mrm{ini}}(\bm{w} \mid \beta)\ln Z(\theta_{\mrm{ini}}). 
\label{eqn:free_energy}
\end{align}
The LC can be obtained from the free energy as
\begin{align}
\chi(\beta)\propto -\frac{\partial^2 f(\beta)}{\partial b \partial c}.
\label{eqn:f_chi_relation}
\end{align}
The evaluation of the free energy is therefore essential. 
The free-energy evaluation will be discussed in the next section. 
The right-hand side of equation (\ref{eqn:f_chi_relation}) is rewritten as
\begin{align*}
-\frac{\partial^2 f(\beta)}{\partial b \partial c}
 = \frac{\partial}{\partial b}\Big(\sum_{j \in H} \int_{-\infty}^{+\infty}d\bm{w}\, P_{\mrm{ini}}(\bm{w} \mid \beta) \mathbb{E}_{\mrm{ini}}[h_j]\Big).
\end{align*}
This expression allows the LC to be viewed as the response of the hidden layer (i.e., the expectations of hidden variables) 
to the stimulus to the visible layer (i.e., the change of visible bias). 
The LC can be understood as a magnetic susceptibility. 
Reference~\cite{Mastromatteo2011} reported that 
learned models tend to concentrate around a critical point in which Fisher information (which is equivalent to the susceptibility) is large 
and models near the critical point are highly sensitive to its learning parameters in the perspective of distinguishability. 
The former result can support the motivation of our hypothesis from the perspective of the goal of learning 
(initial models at the critical point seem reasonable because learned models are expected to be close to the critical point). 
The latter implies that learning models at the critical point are the most sensitive to changes in the learning parameters, 
which supports the reasonability that the susceptibility is maximized in the initial RBM. 

\section{Statistical Mechanical Analysis}
\label{sec:statistical_mechanical_analysis}

The free energy and layer-correlation in equations (\ref{eqn:free_energy}) and (\ref{eqn:layer_correlation}), respectively, 
are evaluated based on the replica method~\cite{Mezard1987,Nishimori2001} which has been developed in the field of statistical mechanics. 
For the evaluation, we assume that the sizes of both layers, $n$ and $m$, are sufficiently large; 
the ratio of both, $\alpha = m / n$, is fixed; the magnitude of $\alpha$ is $O(1)$ for the layer sizes.
The free energy in equation (\ref{eqn:free_energy}) is rewritten as
\begin{align}
f(\beta)= - \frac{1}{n + m}\lim_{x \to 0} \frac{\Phi_x(\beta) - 1}{x},
\label{eqn:replica_relation}
\end{align}
where 
\begin{align}
\Phi_x(\beta):= \int_{-\infty}^{+\infty}d\bm{w}\,P_{\mrm{ini}}(\bm{w} \mid \beta) Z(\theta_{\mrm{ini}})^x.
\label{eqn:Phi_x}
\end{align}
In the replica method, $\Phi_x(\beta)$ is evaluated in which $x$ is limited to natural numbers. 
Subsequently, $x$ is extended to real numbers via analytic continuation, 
and the limit of $x \to 0$ is considered for the evaluation result to obtain the free energy according to equation (\ref{eqn:replica_relation}) 
(this is the so-called \textit{replica trick}). 

\subsection{Free energy evaluation based on replica method}
\label{sec:free_energy_analysis}

Using the replica method with the replica-symmetric (RS) assumption~\cite{Mezard1987,Nishimori2001} , for $n,m \to +\infty$, 
equation (\ref{eqn:Phi_x}) can be expressed as
\begin{align}
\Phi_x(\beta)&=\varepsilon(x)
\exp xn \Big[-\frac{\alpha\beta^2}{2(1 + \alpha)} q_{\mrm{v}}q_{\mrm{h}}
+\frac{1}{2} \hat{q}_{\mrm{v}} (q_{\mrm{v}}-1)
+\frac{\alpha}{2} \hat{q}_{\mrm{h}}q_{\mrm{h}}
+ \int_{-\infty}^{+\infty} Dz \ln \sum_{v \in \mcal{I}}\exp\big(-E_{\mrm{v}}(v,z)\big)
\nn
\aleq
+ \alpha \int_{-\infty}^{+\infty} Dz \ln \sum_{h \in \mcal{X}_{\mrm{h}}}\exp\big(-E_{\mrm{h}}(h,z)\big) + O(x)\Big],
\label{eqn:Phi_x-RS-HS}
\end{align}
where $\varepsilon(x):= (2\pi)^{-x(x-1)}$. Here, 
\begin{align*}
E_{\mrm{v}}(v,z)&:= -\big(b + z \sqrt{\hat{q}_{\mrm{v}}} \big)v, \nn
E_{\mrm{h}}(h,z)&:= -\big(c + z \sqrt{\hat{q}_{\mrm{h}}} \big)h
- \frac{1}{2}\Big(\frac{\beta^2}{1 + \alpha} - \hat{q}_{\mrm{h}} \Big)h^2
\end{align*}
are the effective energies, and 
\begin{align*}
\int_{-\infty}^{+\infty} Dz := \frac{1}{\sqrt{2\pi}}\int_{-\infty}^{+\infty} dz \exp\Big(-\frac{z^2}{2}\Big)
\end{align*}
represents the standard Gaussian measure. In this expression, $q_{\mrm{v}}$ and $q_{\mrm{h}}$ are regarded as the order parameters, 
and $\hat{q}_{\mrm{v}}$ and $\hat{q}_{\mrm{h}}$ are the auxiliary parameters. 
The order and auxiliary parameters are the solution to the saddle point equations, i.e., 
they are the solution to the extremum condition of the exponent of equation (\ref{eqn:Phi_x-RS-HS}). 
In equation (\ref{eqn:Phi_x-RS-HS}), $x$ is assumed to be a real number that is negligibly small.
The derivation of equation (\ref{eqn:Phi_x-RS-HS}) is described in Appendix \ref{app:replica_method}.

From equations (\ref{eqn:replica_relation}) and (\ref{eqn:Phi_x-RS-HS}), 
we obtain the free energy expression as
\begin{align}
f(\beta)&\approx \frac{\alpha\beta^2}{2(1 + \alpha)^2} q_{\mrm{v}}q_{\mrm{h}}
-\frac{1}{2(1 + \alpha)} \hat{q}_{\mrm{v}}(q_{\mrm{v}}-1)
-\frac{\alpha}{2(1 + \alpha)} \hat{q}_{\mrm{h}}q_{\mrm{h}}
- \frac{1}{1 + \alpha}\int_{-\infty}^{+\infty} Dz \ln \sum_{v \in \mcal{I}}\exp\big(-E_{\mrm{v}}(v,z)\big)
\nn
\aleq
- \frac{\alpha}{1 + \alpha} \int_{-\infty}^{+\infty} Dz \ln \sum_{h \in \mcal{X}_{\mrm{h}}}\exp\big(-E_{\mrm{h}}(h,z)\big).
\label{eqn:free_energy_RS}
\end{align} 
In equation (\ref{eqn:free_energy_RS}), $q_{\mrm{v}}$, $q_{\mrm{h}}$, $\hat{q}_{\mrm{v}}$, and $\hat{q}_{\mrm{h}}$ 
are the solution to the saddle point equations, i.e., 
the extremum condition of the free energy. 
From the extremum condition of the free energy, the following saddle point equations are obtained.
\begin{align}
\begin{pmatrix}
\hat{q}_{\mrm{v}}\\
\hat{q}_{\mrm{h}}
\end{pmatrix}
=\beta^2\bm{T}_{\alpha} 
\begin{pmatrix}
q_{\mrm{v}}\\
q_{\mrm{h}}
\end{pmatrix}
,\quad \text{where}\quad 
\bm{T}_{\alpha}:=\frac{1}{1 + \alpha}
\begin{pmatrix}
0 & \alpha\\
1 & 0
\end{pmatrix}
,
\label{eqn:saddle_Q-hat}
\end{align}
\begin{align}
q_{\mrm{v}} =\int_{-\infty}^{+\infty} Dz\, \mathbb{E}_{\mrm{v}}[v \mid z]^2 =\int_{-\infty}^{+\infty} Dz \tanh^2\big(b + z \sqrt{\hat{q}_{\mrm{v}}}\big),
\label{eqn:saddle_Qv}
\end{align}
and
\begin{align}
q_{\mrm{h}} &=\int_{-\infty}^{+\infty} Dz\, \mathbb{E}_{\mrm{h}}[h \mid z]^2
= 
\begin{dcases}
\int_{-\infty}^{+\infty} Dz \tanh^2\big(c + z \sqrt{\hat{q}_{\mrm{h}}}\big),& \mcal{X}_{\mrm{h}} = \mcal{I}\\
\int_{-\infty}^{+\infty} Dz \sig\Big(c  + \frac{\beta^2}{2(1 + \alpha)}-\frac{\hat{q}_{\mrm{h}}}{2}+ z \sqrt{\hat{q}_{\mrm{h}}}\Big)^2,& \mcal{X}_{\mrm{h}} = \mcal{B}
\end{dcases}
.
\label{eqn:saddle_Qh}
\end{align}
Here, $\sig(x) := 1 / (1 + e^{-x})$ is the sigmoid function and
\begin{align*}
\mathbb{E}_{\mrm{v}}[\cdots \mid z]&:= \frac{\sum_{v \in \mcal{I}}(\cdots)\exp(-E_{\mrm{v}}(v,z))}
{\sum_{v \in \mcal{I}}\exp(-E_{\mrm{v}}(v,z))},\nn
\mathbb{E}_{\mrm{h}}[\cdots \mid z]&:= \frac{\sum_{h \in \mcal{X}_{\mrm{h}}}(\cdots)\exp(-E_{\mrm{h}}(h,z))}
{\sum_{h \in \mcal{X}_{\mrm{h}}}\exp(-E_{\mrm{h}}(h,z))}.
\end{align*}
A partial integration below was used in the derivations of equations (\ref{eqn:saddle_Qv}) and (\ref{eqn:saddle_Qh}).
For function $f(z)$ that satisfies $\lim_{z \to \pm \infty} f(z) \exp(-z^2/2) = 0$, 
because $\partial \exp(-z^2/2)/ \partial z = - z \exp(-z^2/2)$, 
\begin{align}
\int_{-\infty}^{+\infty} Dz\, z f(z) =  \int_{-\infty}^{+\infty} Dz\, \frac{\partial  f(z)}{\partial z}.
\label{eqn:partial_integration}
\end{align}
is obtained using the partial integration. 
In addition, $h^2=1$ when $\mcal{X}_{\mrm{h}} = \mcal{I}$ and $h^2 = h$ when $\mcal{X}_{\mrm{h}} = \mcal{B}$ were used in the derivation of the second equation in equation (\ref{eqn:saddle_Qh}). 
When $\mcal{X}_{\mrm{h}} = \mcal{I}$, 
the free energy expression in equation (\ref{eqn:free_energy_RS}) is essentially the same as the ones obtained in references~\cite{Barra2011,Barra2017,Hartnett2018}. 
The saddle point equations (\ref{eqn:saddle_Q-hat}), (\ref{eqn:saddle_Qv}), and (\ref{eqn:saddle_Qh}) can be numerically solved 
using a successive substitution or Newton methods. 

When $\mcal{X}_{\mrm{h}} = \mcal{I}$ and $b = c = 0$, the saddle point equations always have a trivial solution, 
that is $q_{\mrm{v}}=q_{\mrm{h}}= \hat{q}_{\mrm{v}} = \hat{q}_{\mrm{h}} = 0$. 
A specific critical point, $\beta = \beta_{\mrm{critical}}$, exists such that
the saddle point equations have only the trivial solution when $\beta < \beta_{\mrm{critical}}$ (called the paramagnetic phase)
and have not only the trivial solution but also a nontrivial positive solution when $\beta > \beta_{\mrm{critical}}$ (called the spin-glass phase). 
This critical point is called the \textit{spin-glass transition point} (more properly, the paramagnetic and spin-glass phase transition point)~\cite{Hartnett2018}. 
When $\beta > \beta_{\mrm{critical}}$, the nonzero solution is selected. 
The spin-glass transition point is obtained as~\cite{Barra2011,Barra2017}
\begin{align}
\beta_{\mrm{critical}}^2=\sqrt{\alpha} + \frac{1}{\sqrt{\alpha}}.
\label{eqn:spin-glass_transition_point}
\end{align}

\subsection{Layer-correlation evaluation}
\label{sec:susp_analysis}

The LC in equation (\ref{eqn:layer_correlation}) is evaluated according to equation (\ref{eqn:f_chi_relation}). 
In the following, $q_{\mrm{v}}$, $q_{\mrm{h}}$, $\hat{q}_{\mrm{v}}$, and $\hat{q}_{\mrm{h}}$ 
represent the solution to the saddle point equations (\ref{eqn:saddle_Q-hat}), (\ref{eqn:saddle_Qv}), and (\ref{eqn:saddle_Qh}). 
The derivatives of the negative of free energy in equation (\ref{eqn:free_energy_RS}) with respect to $b$ and $c$ are
\begin{align*}
M_{\mrm{v}}&:= -\frac{\partial f(\beta)}{\partial b} = \frac{1}{1+\alpha}\int_{-\infty}^{+\infty} Dz\, \mathbb{E}_{\mrm{v}}[v \mid z],\\
M_{\mrm{h}}&:= -\frac{\partial f(\beta)}{\partial c} = \frac{\alpha}{1+\alpha}\int_{-\infty}^{+\infty} Dz\, \mathbb{E}_{\mrm{h}}[h \mid z],
\end{align*}
respectively.
From equation (\ref{eqn:f_chi_relation}), the LC can be obtained by $\chi(\beta) \propto \partial M_{\mrm{h}} / \partial b$.
We obtain the LC by evaluating susceptibility matrix $\chi_{\ell, k} := \partial M_{\ell} / \partial d_k$ where $\ell, k \in \{\mrm{v}, \mrm{h}\}$; 
here, $d_{\mrm{v}} = b$ and $d_{\mrm{h}} = c$.

For $\ell,k \in \{\mrm{v}, \mrm{h}\}$, the following equation is obtained. 
\begin{align}
\chi_{\ell, k}&= \frac{\tau_{\ell}}{1 + \alpha} \delta(\ell, k) \int_{-\infty}^{+\infty} Dz\, \big\{E_{\ell}^{(2)}(z) - E_{\ell}^{(1)}(z)^2\big\}
- \frac{\tau_{\ell}}{1 + \alpha}\hat{Q}_{\ell, k}\int_{-\infty}^{+\infty} Dz\,  E_{\ell}^{(1)}(z)\big\{ E_{\ell}^{(2)}(z)- E_{\ell}^{(1)}(z)^2\big\},
\label{eqn:chi_(l,k)}
\end{align}
where $\tau_{\ell} =\delta(\mrm{v}, \ell) +  \alpha \delta(\mrm{h}, \ell)$,
\begin{align}
E_{\ell}^{(r)}(z):=
\begin{dcases}
\mathbb{E}_{\mrm{v}}[v^r \mid z],& \ell = \mrm{v}\\
\mathbb{E}_{\mrm{h}}[h^r \mid z],& \ell = \mrm{h}
\end{dcases}
,
\label{eqn:def_E(z)}
\end{align}
and matrix $\hat{Q}_{\ell, k}$ is defined by $\hat{Q}_{\ell, k} := \partial \hat{q}_{\ell} / \partial d_k$. 
Here, $\delta(\ell, k)$ represents the Kronecker delta function. 
Matrix $Q_{\ell, k} := \partial q_{\ell} / \partial d_k$ is obtained as
\begin{align}
Q_{\ell, k}&=2\delta(\ell, k) \int_{-\infty}^{+\infty} Dz\, E_{\ell}^{(1)}(z)\big\{E_{\ell}^{(2)}(z) - E_{\ell}^{(1)}(z)^2\big\}\nn
\aleq
+ \hat{Q}_{\ell, k}\int_{-\infty}^{+\infty} Dz\, \big\{E_{\ell}^{(2)}(z)^2 - 4E_{\ell}^{(2)}(z) E_{\ell}^{(1)}(z)^2 
+ 3 E_{\ell}^{(1)}(z)^4\big\}.
\label{eqn:Q_(l,k)}
\end{align}
The details of the derivation of equations (\ref{eqn:chi_(l,k)}) and (\ref{eqn:Q_(l,k)}) are described in Appendix \ref{app:derivation_chi&Q}.
From equations (\ref{eqn:chi_(l,k)}) and (\ref{eqn:Q_(l,k)}), the matrix expression of the susceptibility matrix can be obtained as
\begin{align}
\bm{\chi}&=\hat{\bm{T}}_{\alpha} \bm{V} - \hat{\bm{T}}_{\alpha}\bm{U}\hat{\bm{Q}},
\label{eqn:mat_chi}\\
\bm{Q}&=2\bm{U} + \bm{W}\hat{\bm{Q}},
\label{eqn:mat_Q}
\end{align}
where $\bm{\chi}$, $\bm{Q}$, $\hat{\bm{Q}}$, and $\hat{\bm{T}}_{\alpha}$ are $2 \times 2$ matrices defined by
\begin{align*}
\bm{\chi}:=
\begin{pmatrix}
\chi_{\mrm{v}, \mrm{v}} & \chi_{\mrm{v}, \mrm{h}}\\
\chi_{\mrm{h}, \mrm{v}} & \chi_{\mrm{h}, \mrm{h}}
\end{pmatrix}
,\quad
\bm{Q}:=
\begin{pmatrix}
Q_{\mrm{v}, \mrm{v}} & Q_{\mrm{v}, \mrm{h}}\\
Q_{\mrm{h}, \mrm{v}} & Q_{\mrm{h}, \mrm{h}}
\end{pmatrix}
,\quad
\hat{\bm{Q}}:=
\begin{pmatrix}
\hat{Q}_{\mrm{v}, \mrm{v}} & \hat{Q}_{\mrm{v}, \mrm{h}}\\
\hat{Q}_{\mrm{h}, \mrm{v}} & \hat{Q}_{\mrm{h}, \mrm{h}}
\end{pmatrix}
,
\end{align*}
and
\begin{align*}
\hat{\bm{T}}_{\alpha}:=\frac{1}{1 + \alpha}
\begin{pmatrix}
1 & 0\\
0 & \alpha
\end{pmatrix}
,
\end{align*}
respectively, and $\bm{V}$, $\bm{U}$, and $\bm{W}$ are $2 \times 2$ diagonal matrices defined by
\begin{align*}
\bm{V}:=
\begin{pmatrix}
V_{\mrm{v}} & 0\\
0 & V_{\mrm{h}}
\end{pmatrix}
,\quad
\bm{U}:=
\begin{pmatrix}
U_{\mrm{v}} & 0\\
0 & U_{\mrm{h}}
\end{pmatrix}
,\quad
\text{and}\quad
\bm{W}:=
\begin{pmatrix}
W_{\mrm{v}} & 0\\
0 & W_{\mrm{h}}
\end{pmatrix}
,\quad
\end{align*}
respectively, where the elements of these diagonal matrices are defined by
\begin{align*}
V_{\ell}&:=\int_{-\infty}^{+\infty} Dz\, \big\{E_{\ell}^{(2)}(z) - E_{\ell}^{(1)}(z)^2\big\},\nn
U_{\ell}&:=\int_{-\infty}^{+\infty} Dz\,  E_{\ell}^{(1)}(z)\big\{ E_{\ell}^{(2)}(z)- E_{\ell}^{(1)}(z)^2\big\},\nn
W_{\ell}&:=\int_{-\infty}^{+\infty} Dz\, \big\{E_{\ell}^{(2)}(z)^2 - 4E_{\ell}^{(2)}(z) E_{\ell}^{(1)}(z)^2 
+ 3 E_{\ell}^{(1)}(z)^4\big\},\nonumber
\end{align*}
for $\ell \in \{\mrm{v}, \mrm{h}\}$. From equation (\ref{eqn:saddle_Q-hat}), 
\begin{align}
\hat{\bm{Q}} = \beta^2 \bm{T}_{\alpha} \bm{Q}.
\label{eqn:mat_Q-hat}
\end{align}
From equations (\ref{eqn:mat_chi}), (\ref{eqn:mat_Q}), and (\ref{eqn:mat_Q-hat}), we obtain the expression of the susceptibility matrix as
\begin{align}
\bm{\chi}=\hat{\bm{T}}_{\alpha} \big\{\bm{V} - 2\beta^2 \bm{U}\bm{T}_{\alpha}
\big(\bm{I} - \beta^2 \bm{W} \bm{T}_{\alpha}\big)^{-1}\bm{U}\big\},
\label{eqn:chi_solution}
\end{align}
where $\bm{I}$ represents a $2 \times 2$ identity matrix.
The LC corresponds to the off-diagonal elements of the susceptibility matrix, i.e., $\chi(\beta) \propto \chi_{\mrm{v}, \mrm{h}}$. 
The susceptibility matrix is numerically obtained by substituting the solution to the saddle point equations (\ref{eqn:saddle_Q-hat}), (\ref{eqn:saddle_Qv}), 
and (\ref{eqn:saddle_Qh}) in equation (\ref{eqn:chi_solution}).

\begin{figure*}[t]
\centering
\includegraphics[height=3.9cm]{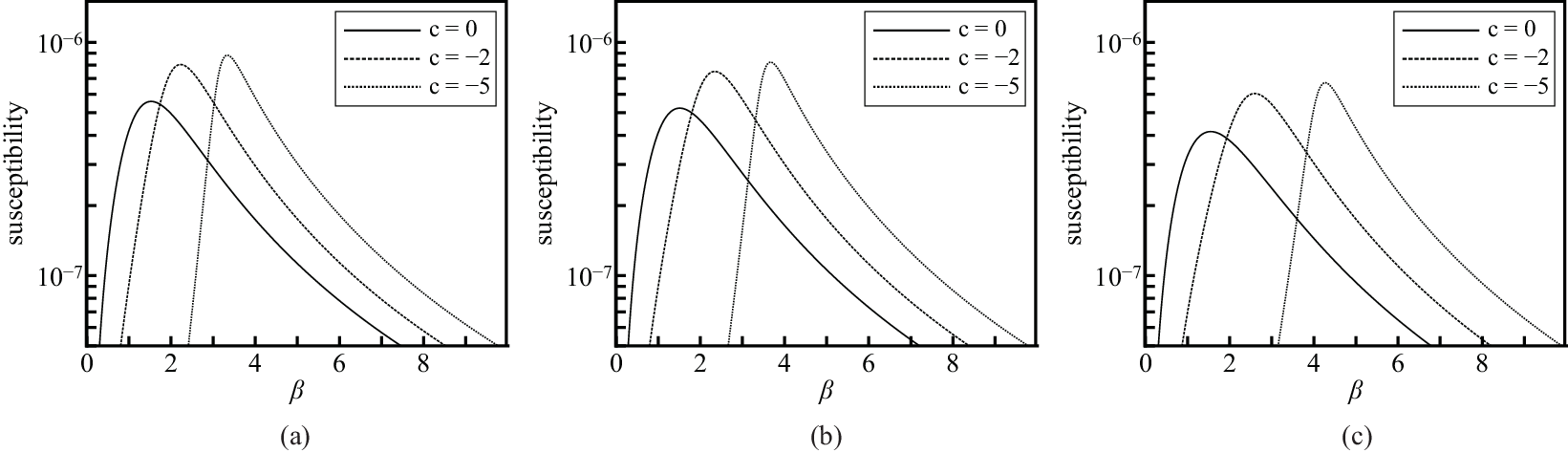}
\caption{Dependency of $|\chi_{\mrm{v}, \mrm{h}}|$ on $\beta$ when $\mcal{X}_{\mrm{h}} = \mcal{B}$: (a) $\alpha = 0.5$, (b) $\alpha = 1$, and (c) $\alpha = 2$.}
\label{fig:susp_binary}
\end{figure*}

\begin{figure*}[t]
\centering
\includegraphics[height=4cm]{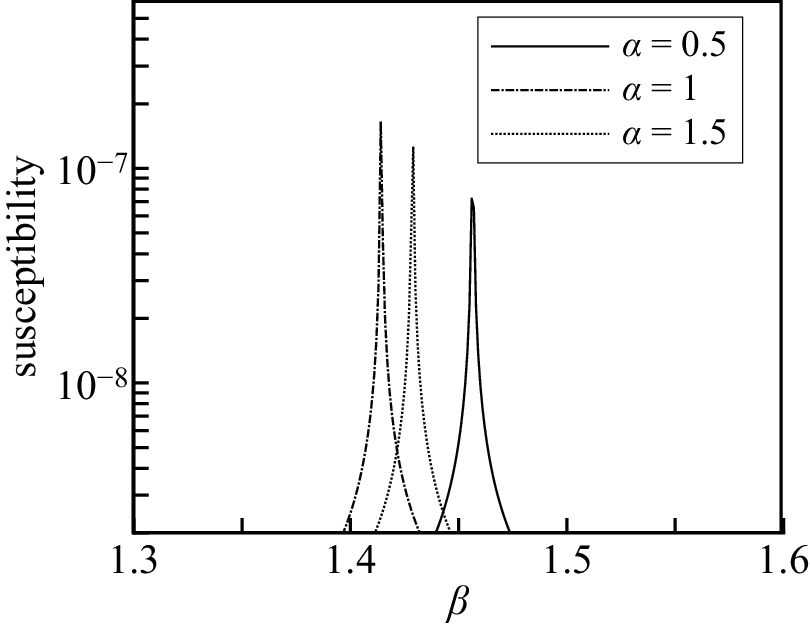}
\caption{Dependency of $|\chi_{\mrm{v}, \mrm{h}}|$ on $\beta$ when $\mcal{X}_{\mrm{h}} =\mcal{I}$.}
\label{fig:susp_ising}
\end{figure*}

The objective is to find the value of $\beta$ that maximizes the absolute value of the LC.
Since $\chi(\beta) \propto \chi_{\mrm{v}, \mrm{h}}$, from equation (\ref{eqn:def_beta_max}), the desired value is obtained by 
\begin{align}
\beta_{\mrm{max}} = \argmax_{\beta} |\chi_{\mrm{v}, \mrm{h}}|.
\label{eqn:beta_max}
\end{align}
The dependency of $|\chi_{\mrm{v}, \mrm{h}}|$ on $\beta$ is demonstrated using numerical experiments. 
Figure \ref{fig:susp_binary} depicts the dependency of $|\chi_{\mrm{v}, \mrm{h}}|$ on $\beta$ when $\mcal{X}_{\mrm{h}} = \mcal{B}$ 
for $\alpha = 0.5, \,1,\, 2$ and $c = 0,\, -5$, in which $b= 0$ is fixed.
Figure \ref{fig:susp_ising} depicts the dependency of $|\chi_{\mrm{v}, \mrm{h}}|$ on $\beta$ when $\mcal{X}_{\mrm{h}} = \mcal{I}$ for 
$\alpha = 0.5,\, 1,\, 1.5$, in which $b = c = 0$ are fixed.
Figures \ref{fig:susp_binary} and \ref{fig:susp_ising} show that $|\chi_{\mrm{v}, \mrm{h}}|$ has the unique maximum at the specific finite value of $\beta$. 
The values of $\beta_{\mrm{max}}$ for various $\alpha$s and $c$s are listed in Appendix \ref{app:list_beta_max}
\footnote{ 
In the experiments demonstrated here (and the experiments to obtain the results listed in table \ref{tab:beta_max_ising} in Appendix \ref{app:list_beta_max}), 
when $\mcal{X}_{\mrm{h}} = \mcal{I}$, the values of $|\chi_{\mrm{v}, \mrm{h}}|$ obtained 
when $b = c = 0$ were very small and buried in numerical errors (e.g., error of numerical integration). 
For example, in this case, $\bm{\chi} = \bm{0}$ when $\beta < \beta_{\mrm{crirical}}$ because $q_{\mrm{v}}=q_{\mrm{h}}= \hat{q}_{\mrm{v}} = \hat{q}_{\mrm{h}} = 0$.
Therefore, we used a very small positive value $\varepsilon \ll 1$ instead of zero, i.e., $b = c = 0$ means $b = c = \varepsilon$, to emphasize the behavior around $\beta = \beta_{\mrm{max}} $. 
Whereas, in other cases, i.e., the cases with $\mcal{X}_{\mrm{h}} = \mcal{B}$, true zero was used. 
}.

In the following, we consider the case where $\mcal{X}_{\mrm{h}} = \mcal{I}$ and $b = c = 0$. 
In this case, a singular behavior of the susceptibility is observed at $\beta = \beta_{\mrm{max}}$, as shown in figure \ref{fig:susp_ising}. 
Such a singular behavior can occur at transition points in many physical systems. 
In fact, in the original SK model~\cite{Sherrington1975}, which has many physical features in common with the bipartite SK model, 
a cusp of the susceptibility appears at the spin-glass transition point.
Moreover, the expression in equation (\ref{eqn:spin-glass_transition_point}) agrees with the numerical results listed in table \ref{tab:beta_max_ising} in Appendix \ref{app:list_beta_max}. 
Therefore, we can conclude 
\begin{align}
\beta_{\mrm{max}}^2  = \beta_{\mrm{critical}}^2=\sqrt{\alpha} + \frac{1}{\sqrt{\alpha}}.
\label{eqn:beta-max-condition_ising}
\end{align}
In a physical interpretation, a model at the transition point is considered unstable, which means that the model can easily change its physical state by a small perturbation. 
Therefore, setting the initial RBM at the transition point seems to be reasonable because such RBM can easily move by a small parameter update.
Notably, from equation (\ref{eqn:beta-max-condition_ising}), $\beta_{\mrm{max}}^2 =2$ when $\alpha = 1$; 
therefore, the proposed weight initialization is identical to that of the Xavier initialization~\cite{Glorot2010} 
when $\mcal{X}_{\mrm{h}} = \mcal{I}$, $b = c = 0$, and $\alpha = 1$ (i.e., when the initial RBM is identical to the Korenblit--Shender model with zero bias~\cite{Korenblit1985}).   

\section{Numerical Experiment}
\label{sec:experiment}

The efficiency of the proposed weight initialization is demonstrated using learning experiments 
based on a toy dataset and three real-world datasets: dry bean (DB)~\cite{DryBean2020}, urban land cover (ULC)~\cite{ULC2013}, and MNIST datasets. 
The efficiency of learning is considered as the growth rate of the log likelihood in equation (\ref{eqn:likelihood}). 
Because the (input) data points in the real-world datasets are not binary, they were binarized to $\mcal{I}$ using the Otsu binarization method~\cite{Otsu1979} 
to adjust to the present RBM in equation (\ref{eqn:RBM}). 
The data-element-wised binarization was applied to the data points in the DB and ULC datasets, i.e., 
$\mrm{v}_i^{(\mu)}$ is binarized in $\{\mrm{v}_i ^{(\mu)}\mid \mu = 1,2,\ldots,N\}$, 
while the data-point-wised binarization was applied to the data points in the MNIST dataset, 
i.e., $\mrm{v}_i^{(\mu)}$ is binarized in $\mbf{v}^{(\mu)}$. 

The learning can be conducted based on a gradient method. 
Although the bias parameters are fixed to constants in the initial RBM, all parameters, including the bias and weight parameters, are tuned during learning.
As mentioned in section \ref{sec:RBM}, the gradients of the log likelihood include the model expectations that require multiple summations over all variables. 
Therefore, the evaluation of the model expectations is infeasible when the size of the RBM is large.
The evaluation of the log likelihood is also infeasible in that case. 
In the experiments for the toy and DB datasets, the gradients and the log likelihoods were exactly evaluated.
Alternatively, in the experiments for the ULC and MNIST datasets, they were evaluated using approximation methods. 
In the evaluation of the model expectations, a layer-wise blocked Gibbs sampling with 1000 sample points was used. 
The details of the sampling are as follows. 
First, the sample points are randomly initialized and relaxed on the initial RBM (the relaxation step is set to 500).
After the relaxation, the transition of the sample points and the update of the learning parameters are repeated 
based on a 40-steps persistent CD (PCD)~\cite{Tieleman2008} (i.e., $\text{PCD}_{40}$ starting from the initial RBM).
In the evaluation of the log likelihood, marginalized annealed importance sampling (mAIS)~\cite{Yasuda2022} was used. 
The mAIS can evaluate the partition function more efficiently than the standard annealed importance sampling. 
In the following experiments, the sample size and annealing schedule size of the mAIS are denoted by $S$ and $K$, respectively. 
In the mAIS, a linear-annealing schedule was employed.

\subsection{Toy dataset}
\label{sec:experiment_toy} 

\begin{figure*}[t]
\centering
\includegraphics[height=2cm]{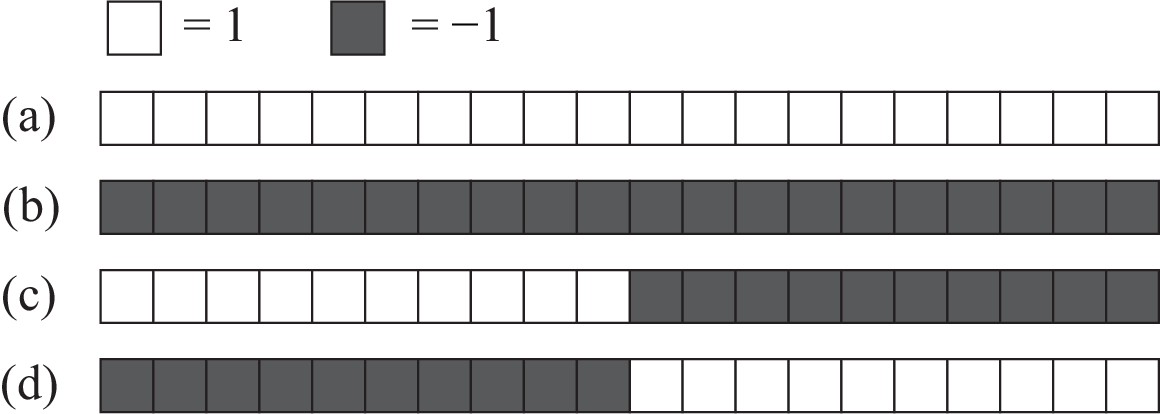}
\caption{Four base patterns with $n = 20$ elements: (a) all are $1$, (b) all are $-1$, (c) the fist 10 elements are $1$ and the others are $-1$, 
and (d) the reverse of (c).}
\label{fig:4BD}
\end{figure*}

\input{table_4BD.tex} 

The learning results using a toy dataset with $n = 20$ and $N = 400$ are presented in this section. 
The toy dataset was obtained by the following procedure. 
First, the four base patterns illustrated in figure \ref{fig:4BD} are prepared. 
Next, from each pattern, 100 data points are generated with $15\%$ randomly flipping. 
A total of $N = 400$ data points are obtained using this procedure.
This toy dataset is similar to that used in reference~\cite{Desjardins2010}. 
For the toy dataset, the RBMs with $n = 20$ and $m = 10,\, 20, \,30$ (i.e., $\alpha = 0.5, \,1, \,1.5$) were used. 
The values of $\beta_{\mrm{max}}$ for each experiment were the corresponding values listed in Appendix \ref{app:list_beta_max}.
For comparison, the experiments when $\beta = \beta_{\mrm{max}}/4, \> \beta_{\mrm{max}}/2, \>2\beta_{\mrm{max}}$, and $4\beta_{\mrm{max}}$ were also conducted.
The results of experiments are listed in Tables \ref{tab:toy_ising_c=0}--\ref{tab:toy_binary_c=-5}. 
The listed values are the average and the standard deviation (in parentheses) over 100 experiments. 
In the learning, batch learning based on the adam optimizer~\cite{Adam2015} with learning rate $\mrm{lr} = 0.01$ was employed. 
The log likelihoods rise with increasing the training epoch. 
After 200 epochs, the result of $\beta = \beta_{\mrm{max}}$ is the best in terms of the log likelihood in all experiments.

\subsection{Real-world datasets: DB, ULC, and MNIST datasets}
\label{sec:experiment_real}

The learning results using real-world datasets, the DB, ULC, and MNIST datasets, are presented in this section. 
For the datasets, the log likelihood with $\beta = \beta_{\mrm{max}}$ is compared with 
the log likelihoods with $\beta = \beta_{\mrm{max}}/4, \> \beta_{\mrm{max}}/2, \>2\beta_{\mrm{max}}$, and $4\beta_{\mrm{max}}$ in several situations, 
as done in the previous toy dataset experiment. 

\begin{figure*}[th]
\centering
\includegraphics[height=4cm]{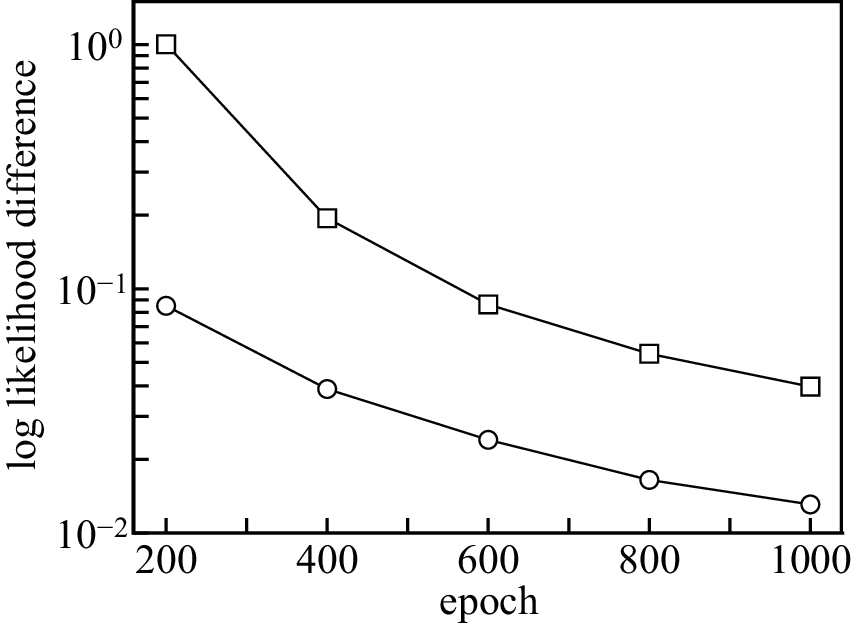}
\caption{Log-likelihood differences on a long-term learning. The square and circle points denote the log likelihood with $\beta = \beta_{\mrm{max}}$ minus 
that with $\beta = 4\beta_{\mrm{max}}$ and with $\beta = \beta_{\mrm{max}}/4$, respectively. 
These points denote the average over 100 experiments.}
\label{fig:DB_likelihood_diff}
\end{figure*}

\input{table_DB.tex} 
In the DB experiment, we used $N=10000$ data points randomly selected from the DB dataset; 
each data point consisted of 16 features. 
For the dataset, the RBMs with $n = 16$ and $m = 16,\, 32$ (i.e., $\alpha = 1, \,2$) were used. 
The values of $\beta_{\mrm{max}}$ for each experiments were the corresponding values listed in Appendix \ref{app:list_beta_max}.
The results of experiments are listed in Tables \ref{tab:db_ising_c=0} and \ref{tab:db_binary}. 
The listed values are the average and the standard deviation (in parentheses) over 150 experiments. 
In the learning, the mini-batch learning based on the adam optimizer with $\mrm{lr} = 0.001$ was employed 
in which the mini-batch size was 500. 
After 200 epochs, the result of $\beta = \beta_{\mrm{max}}$ exhibits the best or at least the second-best performance in the perspective of the growth rate of the log-likelihood. 
Figure \ref{fig:DB_likelihood_diff} depicts the log-likelihood differences (the log likelihood with $\beta = \beta_{\mrm{max}}$ minus 
that with $\beta = \beta_{\mrm{max}}/4$ and with $\beta = 4\beta_{\mrm{max}}$, respectively) on long-term learning in which $\alpha = 1$ and $c = -5$. 
Although the differences become smaller as the training epochs get longer, 
we can observe that the effect of the $\beta$ choice persists over a long time.

\input{table_ULC.tex} 
In the ULC experiment, we used $N=500$ data points randomly selected from the ULC dataset; 
each data point consisted of 147 features. 
For the dataset, the RBM with $n = 147$ and $m = 200$ (i.e., $\alpha \approx 1.36$) was used. 
The results of experiments are listed in Tables \ref{tab:ULC_ising_c=0} and \ref{tab:ULC_binary}. 
The listed values are the average and the standard deviation (in parentheses) over 50 experiments. 
In the learning, the mini-batch learning based on the adam optimizer with $\mrm{lr} = 0.0001$ was employed 
in which the mini-batch size was 50. 
The log likelihood evaluation was conducted using the mAIS with $S=4000$ and $K=2500\>$\footnote{
To check the validity of the setting, we compared with the log likelihood obtained by the mAIS with $S=K=10000$. 
The difference of the log likelihoods was less than $0.1\%$ on average. 
The same verification was conducted in the MNIST experiment.}.
After 100 epochs, the result of $\beta = \beta_{\mrm{max}}$ exhibited the best or at least the second-best performance. 

\input{table_MNIST.tex} 
In the MNIST experiment, we used $N=3000$ data points randomly selected from the MNIST dataset; 
each data point consisted of 784 features. 
For the dataset, the RBM with $n = 784$ and $m = 500$ (i.e., $\alpha \approx 0.64$) was used. 
The results of experiments are listed in Tables \ref{tab:MNIST_ising_c=0} and \ref{tab:MNIST_binary}. 
The listed values are the average and the standard deviation (in parentheses) over 30 experiments. 
In the learning, the mini-batch learning based on the adam optimizer with $\mrm{lr} = 0.0001$ was employed 
in which the mini-batch size was 100. 
The log likelihood evaluation was conducted using the mAIS with $S=4500$ and $K=3000$.
After 100 epochs, the result of $\beta = \beta_{\mrm{max}}$ exhibited the best or at least the second-best performance. 

\section{Summary and Future Studies}
\label{sec:summary}

Based on the statistical mechanical approach, we propose an appropriate standard deviation, $\sigma = \beta / \sqrt{n + m}$, of the Gaussian distribution for the random weight-initialization of Bernoulli--Bernoulli RBMs in which the sample space of the visible variables is $\mcal{I}$ 
and that of the hidden variables is $\mcal{X}_{\mrm{h}}=\mcal{B}$ or $\mcal{X}_{\mrm{h}}=\mcal{I}$. 
In the proposed initialization, the biases of the visible layer are fixed to zero and those of the hidden layer are fixed to a constant $c \leq 0$. 
The appropriate $\beta$ value, i.e., $\beta_{\mrm{max}}$, is defined as the value that maximizes the absolute value of the LC between the visible and hidden layers. 
This criterion is based on the hypothesis discussed in section \ref{sec:RBM}. 
The numerical results presented in section \ref{sec:experiment} support the validity of the proposed initialization. 
The $\beta_{\mrm{max}}$ value depends on only three components: the size ratio between visible and hidden layers $\alpha = m / n$, 
value of $c$, and $\mcal{X}_{\mrm{h}}$. The examples of the $\beta_{\mrm{max}}$ value are listed in Appendix \ref{app:list_beta_max}. 

The four remaining issues that should be addressed in future studies are as follows. 
First, the proposed initialization method does not cover a Gaussian--Bernoulli RBM (GBRBM) that has a continuous visible layer~\cite{Hinton2006b,Cho2011,Takahashi2016,Yasuda2024a} and its variant model~\cite{Yasuda2024b}. 
GBRBMs can treat continuous data, and therefore, they are more important for practical applications. 
To obtain an appropriate dataset-free initialization for GBRBMs need to be extended to the presented framework.
An initialization method for RBMs with binary visible and continuous hidden layers was proposed based on the concept of the Hopfield neural network~\cite{Leonelli2021}. 
This work is different from ours because the initialization method proposed in this work is a ``dataset-oriented'' one 
and our objective model is an RBM with a continuous visible layer; 
however, the results obtained from this work can be a good reference for future studies. 

Second, although our study aims to construct a data-free initialization, an initialization method that uses information in a given dataset may be more practical. 
For example, the visible biases are often initialized using the dataset as $b_i = \tanh^{-1}(N^{-1}\sum_{\mu=1}^N \mrm{v}_i^{(\mu)})$ 
to adjust the expectations of visible variables in the initial RBM to the averages of the dataset, where $\tanh^{-1}(x)$ denotes the inverse function of $\tanh(x)$. 
Unfortunately, however, the presented scheme does not directly cover such cases of heterogeneous initialization of the biases. 
Extending our scheme to cover such cases is also an important future direction.

Third, the explicit expression of $\beta_{\mrm{max}}$ as a function of $\alpha$, $c$, and $\mcal{X}_{\mrm{h}}$ is unknown 
except for the case of $b = c = 0$ and $\mcal{X}_{\mrm{h}}=\mcal{I}$. 
Figures \ref{fig:susp_binary} and \ref{fig:susp_ising} indicate that $|\chi(\beta)|$ has a unique maximum at the specific $\beta$, 
and therefore, $\beta_{\mrm{max}}$ can be obtained numerically using a search method. 
This numerical search is not computationally expensive (it often finishes in a few seconds in a Python implementation) 
but the analytical expression of $\beta_{\mrm{max}}$ is useful for both practical and theoretical aspects. 

Fourth, the proposed initialization is identical to Xavier initialization~\cite{Glorot2010} 
when $\mcal{X}_{\mrm{h}} = \mcal{I}$, $b = c = 0$, and $\alpha = 1$. 
This similarity would suggest some relationship between the concept of Xavier initialization (i.e., conservation of signal variances in forward and backward propagations) 
 and our hypothesis. 
A deeper consideration of the relationship could gain an alternative insight into our hypothesis, increasing the reasonability of our hypothesis.

\section*{Acknowledgment}
This work was partially supported by JSPS KAKENHI (Grant No. 21K11778). 
We would like to thank K. Koyama for the variable discussions.

\appendix
\section{Replica Analysis}
\label{app:replica_method}

For a natural number $x$, performing the Gaussian integral in equation (\ref{eqn:Phi_x}) yields 
\begin{align*}
\Phi_x(\beta) &= \sum_{\bm{v}_{\mrm{all}}}\sum_{\bm{h}_{\mrm{all}}}\exp\Big\{ b\sum_{s=1}^x \sum_{i\in V}v_{i,s}
+ c\sum_{s=1}^x \sum_{j\in H}h_{j,s} 
+ \frac{\beta^2}{2(n + m)} \sum_{i \in V}\sum_{j \in H}\Big(\sum_{s=1}^x v_{i,s}h_{j,s}\Big)^2\Big\},
\end{align*}
where $s = 1,2,\ldots, x$ is the replica index, 
and $\bm{v}_{\mrm{all}} := \{v_{i,s} \in \mcal{I} \mid i \in V;\> s = 1,2,\ldots, x\}$ and
$\bm{h}_{\mrm{all}} := \{h_{j,s} \in \mcal{X}_H \mid j \in H;\> s = 1,2,\ldots, x\}$ 
are the visible and hidden variables in the $x$-replicated system. 
Because $v_{i,s}^2=1$, 
\begin{align}
\Phi_x(\beta) &= \sum_{\bm{v}_{\mrm{all}}}\sum_{\bm{h}_{\mrm{all}}}\exp\Big\{ b\sum_{s=1}^x \sum_{i\in V}v_{i,s}
+ c\sum_{s=1}^x \sum_{j\in H}h_{j,s} 
+ \frac{\beta^2}{n + m}\sum_{s < t}^x \Big(\sum_{i \in V}v_{i,s}v_{i,t}\Big)
\Big(\sum_{j \in H}h_{j,s}h_{j,t}\Big)\nn
\aleq
+ \frac{n\beta^2}{2(n + m)} \sum_{s = 1}^x\sum_{j \in H}h_{j,s}^2 \Big\},
\label{eqn:Phi_x-0}
\end{align}
where $\sum_{s < t}^x$ denotes the sum over all distinct pairs of replicas, i.e., $\sum_{s < t}^x = \sum_{s = 1}^x \sum_{t = s + 1}^x$. 
Using the Dirac delta function, equation (\ref{eqn:Phi_x-0}) is rewritten as
\begin{align*}
\Phi_x(\beta) &= \int_{-\infty}^{+\infty}d\bm{q}^{(\mrm{v})}
 \int_{-\infty}^{+\infty}d\bm{q}^{(\mrm{h})}
\sum_{\bm{v}_{\mrm{all}}}\sum_{\bm{h}_{\mrm{all}}}\exp\Big\{ b\sum_{s=1}^x \sum_{i\in V}v_{i,s}
+ c\sum_{s=1}^x \sum_{j\in H}h_{j,s} 
+ \frac{n m\beta^2}{n + m}\sum_{s < t}^x q_{s,t}^{(\mrm{v})}q_{s,t}^{(\mrm{h})}
\nn
\aleq
+ \frac{n \beta^2}{2(n + m)} \sum_{s = 1}^x\sum_{j \in H}h_{j,s}^2\Big\}
\Big[\prod_{s<t}^x\delta \Big(\sum_{i \in V}v_{i,s}v_{i,t} - n q_{s,t}^{(\mrm{v})} \Big)
\delta \Big(\sum_{j \in H}h_{j,s}h_{j,t} - m q_{s,t}^{(\mrm{h})} \Big)\Big],
\end{align*}
where $\bm{q}^{(\mrm{v})}:= \{ q_{s,t}^{(\mrm{v})} \mid s < t;\> s,t = 1,2,\ldots, x\}$ and 
$\bm{q}^{(\mrm{h})}:= \{ q_{s,t}^{(\mrm{h})} \mid s < t;\> s,t = 1,2,\ldots, x\}$ 
are called \textit{spin-glass order parameters} in statistical mechanics. 
Applying the Fourier integral expression of Dirac delta function,
\begin{align*}
\delta(x) = \frac{1}{2\pi} \int_{-\infty}^{+\infty}d\hat{x} \exp(i\hat{x} x),
\end{align*}
where $i$ is the imaginary number and $\hat{x}$ is the auxiliary parameter, leads to
\begin{align*}
\Phi_x(\beta) &= \varepsilon(x)\int_{-\infty}^{+\infty}d\bm{q}^{(\mrm{v})}
\int_{-\infty}^{+\infty}d\bm{q}^{(\mrm{h})} 
\int_{-\infty}^{+\infty}d\hat{\bm{q}}^{(\mrm{v})}
\int_{-\infty}^{+\infty}d\hat{\bm{q}}^{(\mrm{h})} 
\sum_{\bm{v}_{\mrm{all}}}\sum_{\bm{h}_{\mrm{all}}}\exp\Big\{ b\sum_{s=1}^x \sum_{i\in V}v_{i,s}\nn
\aleq
+ c\sum_{s=1}^x \sum_{j\in H}h_{j,s} + \frac{n m\beta^2}{n + m}\sum_{s < t}^x q_{s,t}^{(\mrm{v})}q_{s,t}^{(\mrm{h})}
+ \frac{n \beta^2}{2(n + m)} \sum_{s = 1}^x \sum_{j \in H}h_{j,s}^2\nn
\aleq
+i\sum_{s < t}^x \hat{q}_{s,t}^{(\mrm{v})}\Big(\sum_{i \in V}v_{i,s}v_{i,t} - n q_{s,t}^{(\mrm{v})} \Big)
+i\sum_{s < t}^x \hat{q}_{s,t}^{(\mrm{h})}\Big(\sum_{j \in H}h_{j,s}h_{j,t} - m q_{s,t}^{(\mrm{h})} \Big)\Big\},
\end{align*}
where $\varepsilon(x)= (2\pi)^{-x(x-1)}$, and $\hat{\bm{q}}^{(\mrm{v})}$ and $\hat{\bm{q}}^{(\mrm{h})}$ are the auxiliary parameters 
of the Fourier integral expression. 
This is rewritten as
\begin{align}
\Phi_x(\beta) 
&= \varepsilon(x)\int_{-\infty}^{+\infty}d\bm{q}^{(\mrm{v})}
\int_{-\infty}^{+\infty}d\bm{q}^{(\mrm{h})}
\int_{-\infty}^{+\infty}d\hat{\bm{q}}^{(\mrm{v})}
\int_{-\infty}^{+\infty}d\hat{\bm{q}}^{(\mrm{h})}
\exp n \Big\{\frac{\alpha\beta^2}{1 + \alpha}\sum_{s < t}^x q_{s,t}^{(\mrm{v})}q_{s,t}^{(\mrm{h})}\nn
\aleq
-i\sum_{s < t}^x \hat{q}_{s,t}^{(\mrm{v})}q_{s,t}^{(\mrm{v})}
-\alpha i\sum_{s < t}^x \hat{q}_{s,t}^{(\mrm{h})}q_{s,t}^{(\mrm{h})}
+ \ln \sum_{\bm{v}^*} \exp\Big(b\sum_{s = 1}^xv_s^* + i\sum_{s < t}^x \hat{q}_{s,t}^{(\mrm{v})}v_{s}^*v_{t}^*\Big)
\nn
\aleq
+ \alpha \ln \sum_{\bm{h}^*} \exp\Big(c\sum_{s = 1}^xh_s^* + \frac{\beta^2}{2(1 + \alpha)}\sum_{s = 1}^x (h_s^*)^2 + i\sum_{s < t}^x \hat{q}_{s,t}^{(\mrm{h})}h_{s}^*h_{t}^*\Big)\Big\},
\label{eqn:Phi_x-3}
\end{align}
where $\bm{v}^*:= \{v_s^* \in \mcal{I} \mid s = 1,2,\ldots, x\}$ and $\bm{h}^*:= \{h_s^* \in \mcal{X}_{\mrm{h}} \mid s = 1,2,\ldots, x\}$.
In the limit of $n \to  +\infty$, the multiple integration in equation (\ref{eqn:Phi_x-3}) can be evaluated by the saddle point method (or the method of steepest descent). 
The saddle point method leads to
\begin{align}
\Phi_x(\beta) &= \varepsilon(x)
\exp n \Big\{\frac{\alpha\beta^2}{1 + \alpha}\sum_{s < t}^x q_{s,t}^{(\mrm{v})}q_{s,t}^{(\mrm{h})}
-\sum_{s < t}^x \hat{q}_{s,t}^{(\mrm{v})}q_{s,t}^{(\mrm{v})}
-\alpha\sum_{s < t}^x \hat{q}_{s,t}^{(\mrm{h})}q_{s,t}^{(\mrm{h})}
+ \ln \sum_{\bm{v}^*} \exp\Big(b\sum_{s = 1}^xv_s^* + \sum_{s < t}^x \hat{q}_{s,t}^{(\mrm{v})}v_{s}^*v_{t}^*\Big)
\nn
\aleq
+ \alpha \ln \sum_{\bm{h}^*} \exp\Big(c\sum_{s = 1}^xh_s^* + \frac{\beta^2}{2(1 + \alpha)}\sum_{s = 1}^x (h_s^*)^2 + \sum_{s < t}^x \hat{q}_{s,t}^{(\mrm{h})}h_{s}^*h_{t}^*\Big)\Big\},
\label{eqn:Phi_x-saddle}
\end{align}
where $\bm{q}^{(\mrm{v})}$, $\bm{q}^{(\mrm{h})}$, $\hat{\bm{q}}^{(\mrm{v})}$, and $\hat{\bm{q}}^{(\mrm{h})}$
represent the saddle points which are the solution to the saddle point equations. 
The saddle point equations are obtained from the extremum condition of the exponent in equation (\ref{eqn:Phi_x-saddle}). 
In equation (\ref{eqn:Phi_x-saddle}), the reparameterizations, $i\hat{q}_{s,t}^{(\mrm{v})} \to \hat{q}_{s,t}^{(\mrm{v})}$ 
and $i\hat{q}_{s,t}^{(\mrm{h})} \to \hat{q}_{s,t}^{(\mrm{h})}$, were used without the loss of generality.

To proceed with the analysis, we employ the RS assumption~\cite{Mezard1987,Nishimori2001} 
that the order and auxiliary parameters do not depend on their index in the saddle point, 
i.e., $q_{s,t}^{(\mrm{v})} = q_{\mrm{v}}$, $q_{s,t}^{(\mrm{h})} = q_{\mrm{h}}$, 
$\hat{q}_{s,t}^{(\mrm{v})} = \hat{q}_{\mrm{v}}$, and $\hat{q}_{s,t}^{(\mrm{h})} = \hat{q}_{\mrm{h}}$ 
hold in the saddle point. With this assumption, equation (\ref{eqn:Phi_x-saddle}) is expressed as
\begin{align}
\Phi_x(\beta)
&=\varepsilon(x)
\exp xn \Big[\frac{(x - 1)\alpha\beta^2}{2(1 + \alpha)} q_{\mrm{v}}q_{\mrm{h}}
+\frac{\hat{q}_{\mrm{v}}}{2}\big( (1-x)q_{\mrm{v}} - 1\big)
-\frac{(x-1)\alpha}{2} \hat{q}_{\mrm{h}}q_{\mrm{h}}\nn
\aleq
+ \frac{1}{x}\ln \sum_{\bm{v}^*} \exp\Big\{b\sum_{s = 1}^xv_s^* + \frac{1}{2}\Big(\sqrt{\hat{q}_{\mrm{v}}}\sum_{s =1}^x v_{s}^*\Big)^2\Big\}\nn
\aleq
+ \frac{\alpha}{x} \ln \sum_{\bm{h}^*} \exp\Big\{c\sum_{s = 1}^xh_s^* + \frac{1}{2}\Big(\frac{\beta^2}{1 + \alpha} - \hat{q}_{\mrm{h}}\Big)\sum_{s = 1}^x (h_s^*)^2 
+ \frac{1}{2}\Big(\sqrt{\hat{q}_{\mrm{h}}}\sum_{s =1}^x h_{s}^*\Big)^2\Big\} \Big].
\label{eqn:Phi_x-RS}
\end{align}
Applying the Hubbard--Stratonovich transformation, 
\begin{align*}
\exp\Big(\frac{\lambda}{2} a^2\Big) = \sqrt{\frac{\lambda}{2 \pi}} \int_{-\infty}^{+\infty}dz \exp\Big(-\frac{\lambda}{2}z^2 + \lambda a  z\Big),
\end{align*}
to equation (\ref{eqn:Phi_x-RS}) provides equation (\ref{eqn:Phi_x-RS-HS}). 
In equation (\ref{eqn:Phi_x-RS-HS}), $x$ was extended to a real number (i.e., replica trick) 
and the expansion, $\ln \int_{-\infty}^{+\infty} Dz  f(z)^x = x \int_{-\infty}^{+\infty} Dz \ln f(z) + O(x^2)$, was applied.

\section{Derivation of Equations (\ref{eqn:chi_(l,k)}) and (\ref{eqn:Q_(l,k)})}
\label{app:derivation_chi&Q}

We consider 
\begin{align*}
I_{\ell,k}^{(r)} := \frac{\partial }{\partial d_k}\int_{-\infty}^{+\infty}Dz E_{\ell}^{(1)}(z)^r,
\end{align*}
where $\ell,k \in \{\mrm{v}, \mrm{h}\}$ and $r \in \{1,2\}$. 
Here, $E_{\ell}^{(r)}(z)$ is defined in equation (\ref{eqn:def_E(z)}), 
and $d_{\mrm{v}} = b$ and  $d_{\mrm{h}} = c$.
$\chi_{\ell,k}$ and $Q_{\ell , k}$ can be expressed in terms of $I_{\ell, k}^{(r)}$ as 
$\chi_{\ell,k}= \tau_{\ell}I_{\ell,k}^{(1)} /(1 + \alpha)$ and $Q_{\ell,k}= I_{\ell,k}^{(2)}$.
\begin{align}
I_{\ell,k}^{(r)} &= r \int_{-\infty}^{+\infty}Dz E_{\ell}^{(1)}(z)^{r-1} \frac{\partial }{\partial d_k} E_{\ell}^{(1)}(z)\nn
&=r\Big( \delta(\ell,k) -\frac{\omega_{\ell}}{2}\hat{Q}_{\ell, k}\Big)\int_{-\infty}^{+\infty}Dz E_{\ell}^{(1)}(z)^{r-1} 
\Big(E_{\ell}^{(2)}(z)  - E_{\ell}^{(1)}(z)^2 \Big)\nn
\aleq
+\frac{r}{2\sqrt{\hat{q}_{\ell}}}\hat{Q}_{\ell, k}
 \int_{-\infty}^{+\infty}Dz\, zE_{\ell}^{(1)}(z)^{r-1} 
\Big(E_{\ell}^{(2)}(z)  - E_{\ell}^{(1)}(z)^2 \Big),
\label{eqn:I_{l.k}-0}
\end{align}
where $\omega_{\ell} = 1$ when $\ell = \mrm{h}$ with $\mcal{X}_{\mrm{h}}=\mcal{B}$, and otherwise, $\omega_{\ell} = 0$. 
Applying equation (\ref{eqn:partial_integration}) to the second term of equation (\ref{eqn:I_{l.k}-0}) yields
\begin{align}
I_{\ell,k}^{(r)}
&=r\Big( \delta(\ell,k) -\frac{\omega_{\ell}}{2}\hat{Q}_{\ell, k}\Big)\int_{-\infty}^{+\infty}Dz E_{\ell}^{(1)}(z)^{r-1} 
\Big(E_{\ell}^{(2)}(z)  - E_{\ell}^{(1)}(z)^2 \Big)\nn
\aleq
+\frac{r(r-1)}{2}\hat{Q}_{\ell, k} \int_{-\infty}^{+\infty}Dz E_{\ell}^{(1)}(z)^{r-2} 
\Big(E_{\ell}^{(2)}(z)  - E_{\ell}^{(1)}(z)^2 \Big)^2\nn
\aleq
+\frac{r}{2}\hat{Q}_{\ell, k} \int_{-\infty}^{+\infty}Dz E_{\ell}^{(1)}(z)^{r-1} 
\Big(E_{\ell}^{(3)}(z)  - E_{\ell}^{(1)}(z)E_{\ell}^{(2)}(z) \Big)\nn
\aleq
+r\hat{Q}_{\ell, k} \int_{-\infty}^{+\infty}Dz E_{\ell}^{(1)}(z)^{r-1} 
\Big(E_{\ell}^{(1)}(z)^3  - E_{\ell}^{(1)}(z)E_{\ell}^{(2)}(z) \Big).
\label{eqn:I_{l.k}-1}
\end{align}
For $t = 1,2,\ldots$, $E_{\ell}^{(2t)}(z) = 1$ and $E_{\ell}^{(2t-1)}(z)=E_{\ell}^{(1)}(z)$ 
when $\ell = \mrm{v}$ or $\ell = \mrm{h}$ with $\mcal{X}_{\mrm{h}}=\mcal{I}$, 
and $E_{\ell}^{(t)}(z)=E_{\ell}^{(1)}(z)$ when $\ell = \mrm{h}$ with $\mcal{X}_{\mrm{h}}=\mcal{B}$. 
This can reduce equation (\ref{eqn:I_{l.k}-1}) to 
\begin{align}
I_{\ell,k}^{(r)}
&=r\delta(\ell,k) \int_{-\infty}^{+\infty}Dz E_{\ell}^{(1)}(z)^{r-1} 
\Big(E_{\ell}^{(2)}(z)  - E_{\ell}^{(1)}(z)^2 \Big)\nn
\aleq
+\frac{r(r-1)}{2}\hat{Q}_{\ell, k} \int_{-\infty}^{+\infty}Dz E_{\ell}^{(1)}(z)^{r-2} 
\Big(E_{\ell}^{(2)}(z)  - E_{\ell}^{(1)}(z)^2 \Big)^2\nn
\aleq
+r\hat{Q}_{\ell, k} \int_{-\infty}^{+\infty}Dz E_{\ell}^{(1)}(z)^{r-1} 
\Big(E_{\ell}^{(1)}(z)^3  - E_{\ell}^{(1)}(z)E_{\ell}^{(2)}(z) \Big).
\label{eqn:I_{l.k}-2}
\end{align}
Equation (\ref{eqn:I_{l.k}-2}) leads to equations (\ref{eqn:chi_(l,k)}) and (\ref{eqn:Q_(l,k)}).

\section{List of $\beta_{\mrm{max}}$}
\label{app:list_beta_max}

Table \ref{tab:beta_max_binary} lists $\beta_{\mrm{max}}$ when $\mcal{X}_{\mrm{h}} = \mcal{B}$ for various values of $\alpha$ and $c$, in which $b$ is fixed to zero. 
Table \ref{tab:beta_max_ising} lists $\beta_{\mrm{max}}$ when $\mcal{X}_{\mrm{h}} = \mcal{I}$ 
for various values of $\alpha$, in which $b$ and $c$ are fixed to zero. 
The values in these tables were obtained by numerically solving equation (\ref{eqn:beta_max}).

\begin{table}[h]
\caption{List of $\beta_{\mrm{max}}$ for various values of $\alpha$ and $c$ when $\mcal{X}_{\mrm{h}} = \mcal{B}$.}
\label{tab:beta_max_binary}
\centering
\scalebox{0.8}{
\begin{tabular}{cc|ccccccc|}
\cline{3-9}
 &  & \multicolumn{7}{c|}{$c$} \\ \cline{3-9} 
 &  & $0$ & $-1$ & $-2$ & $-3$ & $-4$ & $-5$ & $-6$ \\ \hline
\multicolumn{1}{|c|}{\multirow{12}{*}{$\alpha$}} & 0.25 & 1.597   & 1.873    & 2.216    & 2.558    & 2.887    & 3.196    & 3.488 \\
\multicolumn{1}{|c|}{}                                                  & 0.5 & 1.529    &  1.840    &  2.227    & 2.618     &  2.989    &  3.338    & 3.666 \\
\multicolumn{1}{|c|}{}                                                  & 0.75 & 1.511    &  1.853    &  2.280    & 2.712     & 3.120     &  3.503    & 3.863 \\
\multicolumn{1}{|c|}{}                                                  & 1 &  1.510   &  1.879    &  2.344    &  2.812    &  3.255    &  3.669    & 4.058  \\
\multicolumn{1}{|c|}{}                                                  & 1.25 & 1.517    & 1.911     &  2.409    & 2.911     & 3.385     &  3.829    & 4.245 \\
\multicolumn{1}{|c|}{}                                                  & 1.5 &  1.527   &  1.944    &  2.473    & 3.007     &  3.512    &  3.982    & 4.424  \\
\multicolumn{1}{|c|}{}                                                  & 1.75 &  1.539   &  1.977    &  2.536    & 3.100     &  3.633    &  4.130    & 4.595  \\
\multicolumn{1}{|c|}{}                                                  & 2 &  1.551   &   2.009   &  2.596    &  3.190    &  3.749   &   4.271   & 4.759  \\
\multicolumn{1}{|c|}{}                                                  & 2.25 &  1.564   &  2.040    &  2.654    & 3.275     &  3.861    &  4.406    & 4.917  \\
\multicolumn{1}{|c|}{}                                                  & 2.5 &  1.576   &   2.070   &  2.710    &  3.358    &   3.968   &  4.536    & 5.069   \\
\multicolumn{1}{|c|}{}                                                  & 2.75 &  1.588   &  2.099    &  2.764    & 3.437     &  4.071    &  4.662    & 5.215  \\
\multicolumn{1}{|c|}{}                                                  & 3 &  1.600  &   2.127   &   2.815   &  3.514    &   4.171   &  4.783    & 5.356  \\ \hline
\end{tabular}
}
\end{table}

\begin{table}[h]
\caption{List of $\beta_{\mrm{max}}$ for various values of $\alpha$ when $\mcal{X}_{\mrm{h}} = \mcal{I}$. 
The listed values are in good agreement with the expression in equation (\ref{eqn:beta-max-condition_ising}).}
\label{tab:beta_max_ising}
\centering
\scalebox{0.8}{
\begin{tabular}{|l|cccccc|}
\hline
$\alpha$ & $0.5$ & $1$ & $1.5$ & $2$ & $2.5$ & $3$ \\ \hline
$\beta_{\mrm{max}}$ &$1.456$  & $1.414$ & $1.429$ & $1.456$ & $1.488$ & $1.520$ \\ \hline
\end{tabular}
}
\end{table}

\bibliographystyle{unsrt}
\bibliography{cite}

\end{document}

%% file: table_4BD.tex
\begin{table}[ht]
\caption{Training log-likelihoods versus training epoch when $\mcal{X}_{\mrm{h}} = \mcal{I}$ and $c = 0$ (toy dataset). }
\label{tab:toy_ising_c=0}
\centering
\scalebox{0.7}{
\begin{tabular}{ccccccccccc}
\cline{3-11}
 &  & \multicolumn{3}{c}{$\alpha = 0.5$} & \multicolumn{3}{c}{$\alpha = 1$} & \multicolumn{3}{c}{$\alpha = 1.5$} \\ \cline{3-11} 
 & \multicolumn{1}{c|}{} & \multicolumn{3}{c|}{epoch} & \multicolumn{3}{c|}{epoch} & \multicolumn{3}{c|}{epoch} \\ \cline{3-11} 
 & \multicolumn{1}{c|}{} & 50 & 100 & \multicolumn{1}{c|}{200} & 50 & 100 & \multicolumn{1}{c|}{200} & 50 & 100 & \multicolumn{1}{c|}{200} \\ \hline
\multicolumn{1}{|c|}{\multirow{10}{*}{$\beta$}} 
& \multicolumn{1}{c|}{\multirow{2}{*}{$\beta_{\text{max}}/4$}} 
	&$-9.84$  & $-9.75$ & \multicolumn{1}{c|}{$-9.64$} 
	&$-9.86$  & $-9.73$ & \multicolumn{1}{c|}{$-9.51$} 
	& $-9.88$ & $-9.71$ & \multicolumn{1}{c|}{$-9.44$} \\
\multicolumn{1}{|c|}{} & \multicolumn{1}{c|}{} & (0.02) & (0.01) & \multicolumn{1}{c|}{(0.02)} & (0.02) & (0.02) & \multicolumn{1}{c|}{(0.04)} & (0.02) & (0.03) & \multicolumn{1}{c|}{(0.06)} \\ \cline{2-11} 
\multicolumn{1}{|c|}{} & \multicolumn{1}{c|}{\multirow{2}{*}{$\beta_{\text{max}}/2$}} 
	&$\bf{-9.83}$& $\bf{-9.73}$ & \multicolumn{1}{c|}{$-9.62$} 
	&$-9.81$  & $-9.67$ & \multicolumn{1}{c|}{$-9.45$} 
	&$-9.80$  &$-9.63$  & \multicolumn{1}{c|}{$-9.34$} \\
\multicolumn{1}{|c|}{} & \multicolumn{1}{c|}{} & (0.02) & (0.02) & \multicolumn{1}{c|}{(0.02)} & (0.02) & (0.02) & \multicolumn{1}{c|}{(0.04)} & (0.02) & (0.03) & \multicolumn{1}{c|}{(0.06)} \\ \cline{2-11} 
\multicolumn{1}{|c|}{} & \multicolumn{1}{c|}{\multirow{2}{*}{$\beta_{\text{max}}$}} 
	&$-9.87$  &$\bf{-9.73}$  & \multicolumn{1}{c|}{$\bf{-9.61}$} 
	&$\bf{-9.80}$  &$\bf{-9.62}$  & \multicolumn{1}{c|}{$\bf{-9.42}$} 
	&$\bf{-9.74}$  &$\bf{-9.54}$  & \multicolumn{1}{c|}{$\bf{-9.27}$} \\
\multicolumn{1}{|c|}{} & \multicolumn{1}{c|}{} & (0.04) & (0.02) & \multicolumn{1}{c|}{(0.02)} & (0.03) & (0.03) & \multicolumn{1}{c|}{(0.03)} & (0.03) & (0.03) & \multicolumn{1}{c|}{(0.05)} \\ \cline{2-11} 
\multicolumn{1}{|c|}{} & \multicolumn{1}{c|}{\multirow{2}{*}{$2\beta_{\text{max}}$}} 
	&$-10.82$  &$-10.06$  & \multicolumn{1}{c|}{$-9.72$} 
	&$-10.60$  &$-9.93$  & \multicolumn{1}{c|}{$-9.53$} 
	&$-10.43$  &$-9.75$  & \multicolumn{1}{c|}{$-9.35$} \\
\multicolumn{1}{|c|}{} & \multicolumn{1}{c|}{} & (0.30) & (0.10) & \multicolumn{1}{c|}{(0.04)} & (0.19) & (0.10) & \multicolumn{1}{c|}{(0.05)} & (0.16) & (0.09) & \multicolumn{1}{c|}{(0.06)} \\ \cline{2-11} 
\multicolumn{1}{|c|}{} & \multicolumn{1}{c|}{\multirow{2}{*}{$4\beta_{\text{max}}$}} 
	&$-19.84$  &$-13.89$  & \multicolumn{1}{c|}{$-11.28$} 
	&$-17.08$  & $-13.24$ & \multicolumn{1}{c|}{$-10.95$} 
	&$-16.38$  & $-12.81$ & \multicolumn{1}{c|}{$-10.57$} \\
\multicolumn{1}{|c|}{} & \multicolumn{1}{c|}{} & (2.35) & (0.82) & \multicolumn{1}{c|}{(0.39)} & (1.30) & (0.63) & \multicolumn{1}{c|}{(0.33)} & (0.93) & (0.47) & \multicolumn{1}{c|}{(0.25)} \\ \hline
\end{tabular}
}
\end{table}

\begin{table}[ht]
\caption{Training log-likelihoods versus training epoch when $\mcal{X}_{\mrm{h}} = \mcal{B}$ and $c = 0$ (toy dataset). }
\label{tab:toy_binary_c=0}
\centering
\scalebox{0.7}{
\begin{tabular}{ccccccccccc}
\cline{3-11}
 &  & \multicolumn{3}{c}{$\alpha = 0.5$} & \multicolumn{3}{c}{$\alpha = 1$} & \multicolumn{3}{c}{$\alpha = 1.5$} \\ \cline{3-11} 
 & \multicolumn{1}{c|}{} & \multicolumn{3}{c|}{epoch} & \multicolumn{3}{c|}{epoch} & \multicolumn{3}{c|}{epoch} \\ \cline{3-11} 
 & \multicolumn{1}{c|}{} & 50 & 100 & \multicolumn{1}{c|}{200} & 50 & 100 & \multicolumn{1}{c|}{200} & 50 & 100 & \multicolumn{1}{c|}{200} \\ \hline
\multicolumn{1}{|c|}{\multirow{10}{*}{$\beta$}} 
& \multicolumn{1}{c|}{\multirow{2}{*}{$\beta_{\text{max}}/4$}} 
	&$-10.09$  & $\bf{-9.82}$ & \multicolumn{1}{c|}{$\bf{-9.76}$} 
	&$-9.97$  & $-9.85$ & \multicolumn{1}{c|}{$-9.75$} 
	& $-9.97$ & $-9.87$ & \multicolumn{1}{c|}{$-9.74$} \\
\multicolumn{1}{|c|}{} & \multicolumn{1}{c|}{} & (0.08) & (0.01) & \multicolumn{1}{c|}{(0.01)} & (0.02) & (0.01) & \multicolumn{1}{c|}{(0.02)} & (0.01) & (0.01) & \multicolumn{1}{c|}{(0.02)} \\ \cline{2-11} 
\multicolumn{1}{|c|}{} & \multicolumn{1}{c|}{\multirow{2}{*}{$\beta_{\text{max}}/2$}} 
	&$\bf{-10.03}$& $\bf{-9.82}$ & \multicolumn{1}{c|}{$\bf{-9.76}$} 
	&$\bf{-9.96}$  & $-9.84$ & \multicolumn{1}{c|}{$-9.73$} 
	&$\bf{-9.94}$  &$-9.84$  & \multicolumn{1}{c|}{$-9.71$} \\
\multicolumn{1}{|c|}{} & \multicolumn{1}{c|}{} & (0.07) & (0.01) & \multicolumn{1}{c|}{(0.01)} & (0.02) & (0.01) & \multicolumn{1}{c|}{(0.02)} & (0.01) & (0.01) & \multicolumn{1}{c|}{(0.02)} \\ \cline{2-11} 
\multicolumn{1}{|c|}{} & \multicolumn{1}{c|}{\multirow{2}{*}{$\beta_{\text{max}}$}} 
	&$-10.22$  &$-9.83$  & \multicolumn{1}{c|}{$\bf{-9.76}$} 
	&$-9.99$  &$\bf{-9.82}$  & \multicolumn{1}{c|}{$\bf{-9.71}$} 
	&$\bf{-9.94}$  &$\bf{-9.82}$  & \multicolumn{1}{c|}{$\bf{-9.69}$} \\
\multicolumn{1}{|c|}{} & \multicolumn{1}{c|}{} & (0.20) & (0.03) & \multicolumn{1}{c|}{(0.02)} & (0.05) & (0.02) & \multicolumn{1}{c|}{(0.02)} & (0.03) & (0.02) & \multicolumn{1}{c|}{(0.02)} \\ \cline{2-11} 
\multicolumn{1}{|c|}{} & \multicolumn{1}{c|}{\multirow{2}{*}{$2\beta_{\text{max}}$}} 
	&$-11.57$  &$-10.07$  & \multicolumn{1}{c|}{$-9.82$} 
	&$-10.50$  &$-9.96$  & \multicolumn{1}{c|}{$-9.77$} 
	&$-10.33$  &$-9.94$  & \multicolumn{1}{c|}{$-9.72$} \\
\multicolumn{1}{|c|}{} & \multicolumn{1}{c|}{} & (0.60) & (0.12) & \multicolumn{1}{c|}{(0.03)} & (0.20) & (0.05) & \multicolumn{1}{c|}{(0.04)} & (0.15) & (0.05) & \multicolumn{1}{c|}{(0.04)} \\ \cline{2-11} 
\multicolumn{1}{|c|}{} & \multicolumn{1}{c|}{\multirow{2}{*}{$4\beta_{\text{max}}$}} 
	&$-17.89$  &$-12.79$  & \multicolumn{1}{c|}{$-10.68$} 
	&$-15.75$  & $-11.81$ & \multicolumn{1}{c|}{$-10.57$} 
	&$-14.03$  & $-11.46$ & \multicolumn{1}{c|}{$-10.38$} \\
\multicolumn{1}{|c|}{} & \multicolumn{1}{c|}{} & (1.68) & (0.88) & \multicolumn{1}{c|}{(0.23)} & (1.65) & (0.40) & \multicolumn{1}{c|}{(0.18)} & (0.99) & (0.29) & \multicolumn{1}{c|}{(0.16)} \\ \hline
\end{tabular}
}
\end{table}

\begin{table}[ht]
\caption{Training log-likelihoods versus training epoch when $\mcal{X}_{\mrm{h}} = \mcal{B}$ and $c = -5$ (toy dataset). }
\label{tab:toy_binary_c=-5}
\centering
\scalebox{0.7}{
\begin{tabular}{ccccccccccc}
\cline{3-11}
 &  & \multicolumn{3}{c}{$\alpha = 0.5$} & \multicolumn{3}{c}{$\alpha = 1$} & \multicolumn{3}{c}{$\alpha = 1.5$} \\ \cline{3-11} 
 & \multicolumn{1}{c|}{} & \multicolumn{3}{c|}{epoch} & \multicolumn{3}{c|}{epoch} & \multicolumn{3}{c|}{epoch} \\ \cline{3-11} 
 & \multicolumn{1}{c|}{} & 50 & 100 & \multicolumn{1}{c|}{200} & 50 & 100 & \multicolumn{1}{c|}{200} & 50 & 100 & \multicolumn{1}{c|}{200} \\ \hline
\multicolumn{1}{|c|}{\multirow{10}{*}{$\beta$}} 
& \multicolumn{1}{c|}{\multirow{2}{*}{$\beta_{\text{max}}/4$}} 
	&$\bf{-10.40}$  & $-10.00$ & \multicolumn{1}{c|}{$-9.88$} 
	&$-10.28$  & $-9.94$ & \multicolumn{1}{c|}{$-9.78$} 
	& $-10.29$ & $-9.95$ & \multicolumn{1}{c|}{$-9.76$} \\
\multicolumn{1}{|c|}{} & \multicolumn{1}{c|}{} & (0.32) & (0.20) & \multicolumn{1}{c|}{(0.12)} & (0.08) & (0.03) & \multicolumn{1}{c|}{(0.02)} & (0.03) & (0.03) & \multicolumn{1}{c|}{(0.02)} \\ \cline{2-11} 
\multicolumn{1}{|c|}{} & \multicolumn{1}{c|}{\multirow{2}{*}{$\beta_{\text{max}}/2$}} 
	&$-10.44$& $\bf{-9.97}$ & \multicolumn{1}{c|}{$-9.81$} 
	&$\bf{-10.18}$  & $-9.87$ & \multicolumn{1}{c|}{$-9.68$} 
	&$-10.18$  &$-9.86$  & \multicolumn{1}{c|}{$-9.58$} \\
\multicolumn{1}{|c|}{} & \multicolumn{1}{c|}{} & (0.29) & (0.16) & \multicolumn{1}{c|}{(0.10)} & (0.05) & (0.02) & \multicolumn{1}{c|}{(0.03)} & (0.04) & (0.03) & \multicolumn{1}{c|}{(0.04)} \\ \cline{2-11} 
\multicolumn{1}{|c|}{} & \multicolumn{1}{c|}{\multirow{2}{*}{$\beta_{\text{max}}$}} 
	&$-11.16$  &$-10.05$  & \multicolumn{1}{c|}{$\bf{-9.79}$} 
	&$-10.36$  &$\bf{-9.85}$  & \multicolumn{1}{c|}{$\bf{-9.60}$} 
	&$\bf{-10.17}$  &$\bf{-9.77}$  & \multicolumn{1}{c|}{$\bf{-9.43}$} \\
\multicolumn{1}{|c|}{} & \multicolumn{1}{c|}{} & (0.46) & (0.14) & \multicolumn{1}{c|}{(0.05)} & (0.16) & (0.04) & \multicolumn{1}{c|}{(0.04)} & (0.08) & (0.05) & \multicolumn{1}{c|}{(0.05)} \\ \cline{2-11} 
\multicolumn{1}{|c|}{} & \multicolumn{1}{c|}{\multirow{2}{*}{$2\beta_{\text{max}}$}} 
	&$-16.43$  &$-12.72$  & \multicolumn{1}{c|}{$-10.65$} 
	&$-16.07$  &$-12.13$  & \multicolumn{1}{c|}{$-10.65$} 
	&$-15.68$  &$-12.13$  & \multicolumn{1}{c|}{$-10.65$} \\
\multicolumn{1}{|c|}{} & \multicolumn{1}{c|}{} & (1.39) & (0.92) & \multicolumn{1}{c|}{(0.30)} & (1.54) & (0.59) & \multicolumn{1}{c|}{(0.25)} & (1.23) & (0.45) & \multicolumn{1}{c|}{(0.23)} \\ \cline{2-11} 
\multicolumn{1}{|c|}{} & \multicolumn{1}{c|}{\multirow{2}{*}{$4\beta_{\text{max}}$}} 
	&$-42.18$  &$-29.60$  & \multicolumn{1}{c|}{$-18.70$} 
	&$-49.58$  & $-32.05$ & \multicolumn{1}{c|}{$-19.85$} 
	&$-49.58$  & $-30.75$ & \multicolumn{1}{c|}{$-20.30$} \\
\multicolumn{1}{|c|}{} & \multicolumn{1}{c|}{} & (5.96) & (3.77) & \multicolumn{1}{c|}{(1.83)} & (7.19) & (4.60) & \multicolumn{1}{c|}{(1.58)} & (7.26) & (3.88) & \multicolumn{1}{c|}{(1.33)} \\ \hline
\end{tabular}
}
\end{table}

%% file: table_DB.tex
\begin{table}[ht]
\caption{Training log-likelihoods versus training epoch when $\mcal{X}_{\mrm{h}} = \mcal{I}$ and $c = 0$ (DB dataset). }
\label{tab:db_ising_c=0}
\centering
\scalebox{0.7}{
\begin{tabular}{cccccccc}
\cline{3-8}
 &  & \multicolumn{3}{c}{$\alpha = 1$} & \multicolumn{3}{c}{$\alpha = 2$} \\ \cline{3-8} 
 & \multicolumn{1}{c|}{} & \multicolumn{3}{c|}{epoch} & \multicolumn{3}{c|}{epoch} \\ \cline{3-8} 
 & \multicolumn{1}{c|}{} & 50 & 100 & \multicolumn{1}{c|}{200} & 50 & 100 & \multicolumn{1}{c|}{200} \\ \hline
\multicolumn{1}{|c|}{\multirow{10}{*}{$\beta$}} & \multicolumn{1}{c|}{\multirow{2}{*}{$\beta_{\text{max}}/4$}} 
	&$-4.54$  &$-4.35$  & \multicolumn{1}{c|}{$\bf{-4.25}$} 
	&$-4.36$  &$-4.26$  & \multicolumn{1}{c|}{$-4.22$} \\
\multicolumn{1}{|c|}{} & \multicolumn{1}{c|}{} & (0.03) & (0.01) & \multicolumn{1}{c|}{(0.008)} & (0.01) & (0.004) & \multicolumn{1}{c|}{(0.003)} \\ \cline{2-8} 
\multicolumn{1}{|c|}{} & \multicolumn{1}{c|}{\multirow{2}{*}{$\beta_{\text{max}}/2$}} 
	&$-4.51$  &$\bf{-4.33}$  & \multicolumn{1}{c|}{$\bf{-4.25}$} 
	&$-4.34$  &$\bf{-4.25}$  & \multicolumn{1}{c|}{$\bf{-4.21}$} \\
\multicolumn{1}{|c|}{} & \multicolumn{1}{c|}{} & (0.03) & (0.01) & \multicolumn{1}{c|}{(0.008)} & (0.01) & (0.005) & \multicolumn{1}{c|}{(0.003)} \\ \cline{2-8} 
\multicolumn{1}{|c|}{} & \multicolumn{1}{c|}{\multirow{2}{*}{$\beta_{\text{max}}$}} 
	&$\bf{-4.49}$  &$\bf{-4.33}$  & \multicolumn{1}{c|}{$\bf{-4.25}$} 
	&$\bf{-4.33}$  &$\bf{-4.25}$  & \multicolumn{1}{c|}{$\bf{-4.21}$} \\
\multicolumn{1}{|c|}{} & \multicolumn{1}{c|}{} & (0.03) & (0.02) & \multicolumn{1}{c|}{(0.008)} & (0.01) & (0.005) & \multicolumn{1}{c|}{(0.003)} \\ \cline{2-8} 
\multicolumn{1}{|c|}{} & \multicolumn{1}{c|}{\multirow{2}{*}{$2\beta_{\text{max}}$}} 
	&$-4.60$  &$-4.37$  & \multicolumn{1}{c|}{$-4.26$} 
	&$-4.37$  &$-4.27$  & \multicolumn{1}{c|}{$-4.22$} \\
\multicolumn{1}{|c|}{} & \multicolumn{1}{c|}{} & (0.07) & (0.02) & \multicolumn{1}{c|}{(0.01)} & (0.03) & (0.01) & \multicolumn{1}{c|}{(0.004)} \\ \cline{2-8} 
\multicolumn{1}{|c|}{} & \multicolumn{1}{c|}{\multirow{2}{*}{$4\beta_{\text{max}}$}} 
	&$-5.68$ &$-4.67$  & \multicolumn{1}{c|}{$-4.35$} 
	&$-4.95$  &$-4.44$  & \multicolumn{1}{c|}{$-4.27$} \\
\multicolumn{1}{|c|}{} & \multicolumn{1}{c|}{} & (0.40) & (0.11) & \multicolumn{1}{c|}{(0.03)} & (0.16) & (0.04) & \multicolumn{1}{c|}{(0.01)} \\ \hline
\end{tabular}
}
\end{table}

\begin{table}[ht]
\caption{Training log-likelihoods versus training epoch when $\mcal{X}_{\mrm{h}} = \mcal{B}$ and $c = 0,\, -5$ (DB dataset). }
\label{tab:db_binary}
\centering
\scalebox{0.7}{
\begin{tabular}{cccccccccccccc}
\cline{3-14}
 &  & \multicolumn{6}{c}{$c = 0$} & \multicolumn{6}{c}{$c = -5$} \\ \cline{3-14} 
 &  & \multicolumn{3}{c}{$\alpha = 1$} & \multicolumn{3}{c}{$\alpha = 2$} & \multicolumn{3}{c}{$\alpha = 1$} & \multicolumn{3}{c}{$\alpha = 2$} \\ \cline{3-14} 
 & \multicolumn{1}{c|}{} & \multicolumn{3}{c|}{epoch} & \multicolumn{3}{c|}{epoch} & \multicolumn{3}{c|}{epoch} & \multicolumn{3}{c|}{epoch} \\ \cline{3-14} 
 & \multicolumn{1}{c|}{} & 50 & 100 & \multicolumn{1}{c|}{200} & 50 & 100 & \multicolumn{1}{c|}{200} & 50 & 100 & \multicolumn{1}{c|}{200} & 50 & 100 & \multicolumn{1}{c|}{200} \\ \hline
\multicolumn{1}{|c|}{\multirow{10}{*}{$\beta$}} & \multicolumn{1}{c|}{\multirow{2}{*}{$\beta_{\mathrm{max}}/4$}} 
	&$-5.06$  &$-4.69$  & \multicolumn{1}{c|}{$-4.41$} &$-4.78$  &$-4.51$  & \multicolumn{1}{c|}{$-4.31$} 
	&$-4.89$  &$-4.61$  & \multicolumn{1}{c|}{$-4.39$} &$-4.70$  &$-4.42$  & \multicolumn{1}{c|}{$-4.29$}\\
\multicolumn{1}{|c|}{} & \multicolumn{1}{c|}{} & (0.08) & (0.02) & \multicolumn{1}{c|}{(0.02)} & (0.02) & (0.02) & \multicolumn{1}{c|}{(0.007)}
	& (0.10) & (0.07) & \multicolumn{1}{c|}{(0.05)} & (0.06) & (0.05) & \multicolumn{1}{c|}{(0.02)} \\ \cline{2-14} 
\multicolumn{1}{|c|}{} & \multicolumn{1}{c|}{\multirow{2}{*}{$\beta_{\mathrm{max}}/2$}} 
	&$-4.97$  &$-4.64$  & \multicolumn{1}{c|}{$-4.38$} &$-4.72$  &$-4.44$  & \multicolumn{1}{c|}{$-4.29$}
	&$\bf{-4.66}$  &$\bf{-4.44}$  & \multicolumn{1}{c|}{$-4.32$} &$\bf{-4.42}$  &$\bf{-4.29}$  & \multicolumn{1}{c|}{$\bf{-4.24}$} \\
\multicolumn{1}{|c|}{} & \multicolumn{1}{c|}{} & (0.05) & (0.03) & \multicolumn{1}{c|}{(0.01)} & (0.03) & (0.02) & \multicolumn{1}{c|}{(0.008)}
	& (0.07) & (0.05) & \multicolumn{1}{c|}{(0.02)} & (0.04) & (0.01) & \multicolumn{1}{c|}{(0.006)} \\ \cline{2-14} 
\multicolumn{1}{|c|}{} & \multicolumn{1}{c|}{\multirow{2}{*}{$\beta_{\mathrm{max}}$}} 
	&$\bf{-4.95}$  &$\bf{-4.58}$  & \multicolumn{1}{c|}{$\bf{-4.36}$}  &$\bf{-4.63}$  &$-4.39$  & \multicolumn{1}{c|}{$-4.28$} 
	&$-4.73$  &$\bf{-4.44}$  & \multicolumn{1}{c|}{$\bf{-4.31}$} &$-4.45$  &$-4.30$  & \multicolumn{1}{c|}{$\bf{-4.24}$} \\
\multicolumn{1}{|c|}{} & \multicolumn{1}{c|}{} & (0.08) & (0.04) & \multicolumn{1}{c|}{(0.02)} & (0.05) & (0.02) & \multicolumn{1}{c|}{(0.008)} 
	& (0.11) & (0.04) & \multicolumn{1}{c|}{(0.02)} & (0.04) & (0.01) & \multicolumn{1}{c|}{(0.006)} \\ \cline{2-14} 
\multicolumn{1}{|c|}{} & \multicolumn{1}{c|}{\multirow{2}{*}{$2\beta_{\mathrm{max}}$}} 
	&$-5.05$  &$\bf{-4.58}$  & \multicolumn{1}{c|}{$\bf{-4.36}$} &$-4.64$  &$\bf{-4.38}$  & \multicolumn{1}{c|}{$\bf{-4.27}$} 
	&$-6.01$  &$-4.87$  & \multicolumn{1}{c|}{$-4.43$}  &$-5.25$  &$-4.54$  & \multicolumn{1}{c|}{$-4.31$} \\
\multicolumn{1}{|c|}{} & \multicolumn{1}{c|}{} & (0.14) & (0.06) & \multicolumn{1}{c|}{(0.02)} & (0.09) & (0.03) & \multicolumn{1}{c|}{(0.01)}
	& (0.44) & (0.16) & \multicolumn{1}{c|}{(0.05)} & (0.25) & (0.08) & \multicolumn{1}{c|}{(0.02)} \\ \cline{2-14} 
\multicolumn{1}{|c|}{} & \multicolumn{1}{c|}{\multirow{2}{*}{$4\beta_{\mathrm{max}}$}} 
	&$-6.00$ &$-4.88$  & \multicolumn{1}{c|}{$-4.42$} &$-5.10$  &$-4.49$  & \multicolumn{1}{c|}{$-4.29$}
	&$-16.05$ &$-8.29$  & \multicolumn{1}{c|}{$-5.36$}  &$-12.63$  &$-7.03$  & \multicolumn{1}{c|}{$-4.95$} \\
\multicolumn{1}{|c|}{} & \multicolumn{1}{c|}{} & (0.43) & (0.17) & \multicolumn{1}{c|}{(0.04)} & (0.25) & (0.07) & \multicolumn{1}{c|}{(0.02)} 
	& (3.08) & (1.03) & \multicolumn{1}{c|}{(0.33)} & (2.56) & (0.84) & \multicolumn{1}{c|}{(0.18)} \\ \hline
\end{tabular}
}
\end{table}

%% file: table_ULC.tex
\begin{table}[ht]
\caption{Training log-likelihoods versus training epoch when $\mcal{X}_{\mrm{h}} = \mcal{I}$ and $c = 0$ (ULC dataset).  
$\beta_{\mrm{max}}\approx 1.42$.}
\label{tab:ULC_ising_c=0}
\centering
\scalebox{0.7}{
\begin{tabular}{cc|ccc|}
\cline{3-5}
 &  & \multicolumn{3}{c|}{epoch} \\ \cline{3-5} 
 &  & 40 & 80 & 100 \\ \hline
\multicolumn{1}{|c|}{\multirow{10}{*}{$\beta$}} & \multirow{2}{*}{$\beta_{\text{max}}/4$} 
	&$-52.69$  &$-46.87$  &$-45.29$  \\
\multicolumn{1}{|c|}{} &  & (0.34) & (0.48) & (0.57) \\ \cline{2-5} 
\multicolumn{1}{|c|}{} & \multirow{2}{*}{$\beta_{\text{max}}/2$} 
	&$\bf{-50.15}$  &$-44.80$  &$-43.25$  \\
\multicolumn{1}{|c|}{} &  & (0.28) & (0.50) & (0.60) \\ \cline{2-5} 
\multicolumn{1}{|c|}{} & \multirow{2}{*}{$\beta_{\text{max}}$} 
	&$-51.00$  &$\bf{-44.38}$  &$\bf{-42.70}$  \\
\multicolumn{1}{|c|}{} &  & (0.53) & (0.70) & (0.77) \\ \cline{2-5} 
\multicolumn{1}{|c|}{} & \multirow{2}{*}{$2\beta_{\text{max}}$} 
	&$-75.50$  &$-56.27$  &$-52.20$  \\
\multicolumn{1}{|c|}{} &  & (1.75) & (0.95) & (0.89) \\ \cline{2-5} 
\multicolumn{1}{|c|}{} & \multirow{2}{*}{$4\beta_{\text{max}}$} 
	&$-205.63$  &$-129.06$  &$-112.19$  \\
\multicolumn{1}{|c|}{} &  & (7.48) & (3.26) & (2.40) \\ \hline
\end{tabular}
}
\end{table}

\begin{table}[ht]
\caption{Training log-likelihoods versus training epoch when $\mcal{X}_{\mrm{h}} = \mcal{B}$ (ULC dataset). 
$\beta_{\mrm{max}}\approx 1.52, \,2.70$, and $3.90$ 
for $c = 0,\,-2.5$, and $-5$, respectively.}
\label{tab:ULC_binary}
\centering
\scalebox{0.7}{
\begin{tabular}{ccccccccccc}
\cline{3-11}
 &  & \multicolumn{3}{c}{$c = 0$} & \multicolumn{3}{c}{$c = -2.5$} & \multicolumn{3}{c}{$c = -5$} \\ \cline{3-11} 
 & \multicolumn{1}{c|}{} & \multicolumn{3}{c|}{epoch} & \multicolumn{3}{c|}{epoch} & \multicolumn{3}{c|}{epoch} \\ \cline{3-11} 
 & \multicolumn{1}{c|}{} & 40 & 80 & \multicolumn{1}{c|}{100} & 40 & 80 & \multicolumn{1}{c|}{100} & 40 & 80 & \multicolumn{1}{c|}{100} \\ \hline
\multicolumn{1}{|c|}{\multirow{10}{*}{$\beta$}} 
& \multicolumn{1}{c|}{\multirow{2}{*}{$\beta_{\text{max}}/4$}} 
	&$-69.46$  & $-62.04$ & \multicolumn{1}{c|}{$-60.88$} 
	&$-66.57$  & $-57.28$ & \multicolumn{1}{c|}{$-54.79$} 
	& $-76.99$ & $-59.62$ & \multicolumn{1}{c|}{$-56.42$} \\
\multicolumn{1}{|c|}{} & \multicolumn{1}{c|}{} & (0.24) & (0.44) & \multicolumn{1}{c|}{(0.35)} & (0.28) & (0.37) & \multicolumn{1}{c|}{(0.42)} & (1.17) & (0.70) & \multicolumn{1}{c|}{(0.65)} \\ \cline{2-11} 
\multicolumn{1}{|c|}{} & \multicolumn{1}{c|}{\multirow{2}{*}{$\beta_{\text{max}}/2$}} 
	&$-66.91$& $-59.52$ & \multicolumn{1}{c|}{$-57.94$} 
	&$\bf{-61.52}$  & $\bf{-53.21}$ & \multicolumn{1}{c|}{$\bf{-50.72}$} 
	&$\bf{-66.31}$  &$\bf{-52.81}$  & \multicolumn{1}{c|}{$\bf{-49.73}$} \\
\multicolumn{1}{|c|}{} & \multicolumn{1}{c|}{} & (0.56) & (0.41) & \multicolumn{1}{c|}{(0.48)} & (0.44) & (0.53) & \multicolumn{1}{c|}{(0.58)} & (0.77) & (0.62) & \multicolumn{1}{c|}{(0.52)} \\ \cline{2-11} 
\multicolumn{1}{|c|}{} & \multicolumn{1}{c|}{\multirow{2}{*}{$\beta_{\text{max}}$}} 
	&$\bf{-65.20}$  &$\bf{-57.04}$  & \multicolumn{1}{c|}{$\bf{-54.50}$} 
	&$-64.24$  &$-53.33$  & \multicolumn{1}{c|}{$-50.82$} 
	&$-73.08$  &$-56.48$  & \multicolumn{1}{c|}{$-53.48$} \\
\multicolumn{1}{|c|}{} & \multicolumn{1}{c|}{} & (0.71) & (0.61) & \multicolumn{1}{c|}{(0.61)} & (1.02) & (0.67) & \multicolumn{1}{c|}{(0.60)} & (1.31) & (0.79) & \multicolumn{1}{c|}{(0.79)} \\ \cline{2-11} 
\multicolumn{1}{|c|}{} & \multicolumn{1}{c|}{\multirow{2}{*}{$2\beta_{\text{max}}$}} 
	&$-74.45$  &$-60.74$  & \multicolumn{1}{c|}{$-57.51$} 
	&$-114.81$  &$-80.20$  & \multicolumn{1}{c|}{$-73.31$} 
	&$-174.97$  &$-115.37$  & \multicolumn{1}{c|}{$-100.31$} \\
\multicolumn{1}{|c|}{} & \multicolumn{1}{c|}{} & (1.41) & (0.88) & \multicolumn{1}{c|}{(0.72)} & (3.53) & (1.95) & \multicolumn{1}{c|}{(1.87)} & (7.75) & (3.98) & \multicolumn{1}{c|}{(2.92)} \\ \cline{2-11} 
\multicolumn{1}{|c|}{} & \multicolumn{1}{c|}{\multirow{2}{*}{$4\beta_{\text{max}}$}} 
	&$-153.74$  &$-104.73$  & \multicolumn{1}{c|}{$-94.52$} 
	&$-353.73$  & $-235.89$ & \multicolumn{1}{c|}{$-202.45$} 
	&$-573.87$  & $-424.75$ & \multicolumn{1}{c|}{$-368.38$} \\
\multicolumn{1}{|c|}{} & \multicolumn{1}{c|}{} & (7.33) & (4.00) & \multicolumn{1}{c|}{(3.07)} & (17.02) & (10.59) & \multicolumn{1}{c|}{(9.49)} & (26.33) & (20.12) & \multicolumn{1}{c|}{(18.53)} \\ \hline
\end{tabular}
}
\end{table}

%% file: table_MNIST.tex
\begin{table}[ht]
\caption{Training log-likelihoods versus training epoch when $\mcal{X}_{\mrm{h}} = \mcal{I}$ and $c = 0$ (MNIST dataset). 
$\beta_{\mrm{max}}\approx 1.43$.}
\label{tab:MNIST_ising_c=0}
\centering
\scalebox{0.7}{
\begin{tabular}{cc|ccc|}
\cline{3-5}
 &  & \multicolumn{3}{c|}{epoch} \\ \cline{3-5} 
 &  & 40 & 80 & 100 \\ \hline
\multicolumn{1}{|c|}{\multirow{10}{*}{$\beta$}} & \multirow{2}{*}{$\beta_{\text{max}}/4$} 
	&$-193.59$  &$-177.69$  &$-175.24$  \\
\multicolumn{1}{|c|}{} &  & (3.60) & (2.54) & (2.45) \\ \cline{2-5} 
\multicolumn{1}{|c|}{} & \multirow{2}{*}{$\beta_{\text{max}}/2$} 
	&$-155.09$  &$-143.80$  &$-141.96$  \\
\multicolumn{1}{|c|}{} &  & (1.75) & (1.47) & (1.65) \\ \cline{2-5} 
\multicolumn{1}{|c|}{} & \multirow{2}{*}{$\beta_{\text{max}}$} 
	&$\bf{-140.14}$  &$\bf{-132.12}$  &$\bf{-131.07}$  \\
\multicolumn{1}{|c|}{} &  & (0.94) & (1.54) & (1.70) \\ \cline{2-5} 
\multicolumn{1}{|c|}{} & \multirow{2}{*}{$2\beta_{\text{max}}$} 
	&$-154.27$  &$-143.46$  &$-142.89$  \\
\multicolumn{1}{|c|}{} &  & (1.64) & (1.58) & (1.51) \\ \cline{2-5} 
\multicolumn{1}{|c|}{} & \multirow{2}{*}{$4\beta_{\text{max}}$} 
	&$-217.65$  &$-197.92$  &$-193.90$  \\
\multicolumn{1}{|c|}{} &  & (2.10) & (3.15) & (3.03) \\ \hline
\end{tabular}
}
\end{table}

\begin{table}[ht]
\caption{Training log-likelihoods versus training epoch when $\mcal{X}_{\mrm{h}} = \mcal{B}$ (MNIST dataset). 
$\beta_{\mrm{max}}\approx 1.52, \,2.46$, and $3.42$ 
for $c = 0,\,-2.5$, and $-5$, respectively.}
\label{tab:MNIST_binary}
\centering
\scalebox{0.7}{
\begin{tabular}{ccccccccccc}
\cline{3-11}
 &  & \multicolumn{3}{c}{$c = 0$} & \multicolumn{3}{c}{$c = -2.5$} & \multicolumn{3}{c}{$c = -5$} \\ \cline{3-11} 
 & \multicolumn{1}{c|}{} & \multicolumn{3}{c|}{epoch} & \multicolumn{3}{c|}{epoch} & \multicolumn{3}{c|}{epoch} \\ \cline{3-11} 
 & \multicolumn{1}{c|}{} & 40 & 80 & \multicolumn{1}{c|}{100} & 40 & 80 & \multicolumn{1}{c|}{100} & 40 & 80 & \multicolumn{1}{c|}{100} \\ \hline
\multicolumn{1}{|c|}{\multirow{10}{*}{$\beta$}} 
& \multicolumn{1}{c|}{\multirow{2}{*}{$\beta_{\text{max}}/4$}} 
	&$-198.15$  & $-160.24$ & \multicolumn{1}{c|}{$-151.90$} 
	&$-515.07$  & $-136.05$ & \multicolumn{1}{c|}{$-313.46$} 
	& $-148.19$ & $-130.27$ & \multicolumn{1}{c|}{$-125.60$} \\
\multicolumn{1}{|c|}{} & \multicolumn{1}{c|}{} & (1.30) & (0.82) & \multicolumn{1}{c|}{(0.69)} & (0.61) & (0.84) & \multicolumn{1}{c|}{(0.79)} & (0.71) & (0.81) & \multicolumn{1}{c|}{(0.90)} \\ \cline{2-11} 
\multicolumn{1}{|c|}{} & \multicolumn{1}{c|}{\multirow{2}{*}{$\beta_{\text{max}}/2$}} 
	&$-170.16$& $-147.30$ & \multicolumn{1}{c|}{$-140.82$} 
	&$-142.09$  & $-128.72$ & \multicolumn{1}{c|}{$-124.28$} 
	&$\bf{-134.65}$  &$-121.48$  & \multicolumn{1}{c|}{$-117.03$} \\
\multicolumn{1}{|c|}{} & \multicolumn{1}{c|}{} & (0.94) & (0.64) & \multicolumn{1}{c|}{(0.68)} & (0.76) & (0.94) & \multicolumn{1}{c|}{(1.13)} & (0.89) & (1.09) & \multicolumn{1}{c|}{(0.92)} \\ \cline{2-11} 
\multicolumn{1}{|c|}{} & \multicolumn{1}{c|}{\multirow{2}{*}{$\beta_{\text{max}}$}} 
	&$-152.00$  &$-135.81$  & \multicolumn{1}{c|}{$-131.20$} 
	&$\bf{-136.62}$  &$\bf{-121.93}$  & \multicolumn{1}{c|}{$\bf{-118.32}$} 
	&$-136.91$  &$\bf{-120.60}$  & \multicolumn{1}{c|}{$\bf{-116.11}$} \\
\multicolumn{1}{|c|}{} & \multicolumn{1}{c|}{} & (0.66) & (0.79) & \multicolumn{1}{c|}{(0.80)} & (0.88) & (0.88) & \multicolumn{1}{c|}{(1.01)} & (1.01) & (1.07) & \multicolumn{1}{c|}{(1.27)} \\ \cline{2-11} 
\multicolumn{1}{|c|}{} & \multicolumn{1}{c|}{\multirow{2}{*}{$2\beta_{\text{max}}$}} 
	&$\bf{-150.47}$  &$\bf{-131.36}$  & \multicolumn{1}{c|}{$\bf{-126.03}$} 
	&$-160.29$  &$-136.63$  & \multicolumn{1}{c|}{$-130.23$} 
	&$-181.42$  &$-152.88$  & \multicolumn{1}{c|}{$-145.20$} \\
\multicolumn{1}{|c|}{} & \multicolumn{1}{c|}{} & (0.96) & (1.16) & \multicolumn{1}{c|}{(1.05)} & (1.58) & (1.07) & \multicolumn{1}{c|}{(1.08)} & (1.48) & (1.31) & \multicolumn{1}{c|}{(1.43)} \\ \cline{2-11} 
\multicolumn{1}{|c|}{} & \multicolumn{1}{c|}{\multirow{2}{*}{$4\beta_{\text{max}}$}} 
	&$-183.86$  &$-155.64$  & \multicolumn{1}{c|}{$-147.78$} 
	&$-226.82$  & $-193.89$ & \multicolumn{1}{c|}{$-184.10$} 
	&$-287.76$  & $-229.52$ & \multicolumn{1}{c|}{$-217.94$} \\
\multicolumn{1}{|c|}{} & \multicolumn{1}{c|}{} & (1.93) & (1.28) & \multicolumn{1}{c|}{(1.13)} & (1.86) & (2.42) & \multicolumn{1}{c|}{(2.33)} & (3.98) & (3.16) & \multicolumn{1}{c|}{(3.23)} \\ \hline
\end{tabular}
}
\end{table}